\documentclass[11pt]{article}

\usepackage{acl}

\usepackage{times}
\usepackage{latexsym}
\usepackage{amssymb}
\usepackage{graphicx}
\usepackage{amsmath}
\usepackage{algorithm}
\usepackage{algpseudocode}
\usepackage{booktabs} % For formal tables
\usepackage{xcolor}
\usepackage{todonotes}
\usepackage{array}
\usepackage{subcaption}
\usepackage[T1]{fontenc}
\usepackage{xspace}
\usepackage{graphicx}
\usepackage{subcaption}
\usepackage{wrapfig}

\usepackage[utf8]{inputenc}
\usepackage{tabularx} 
\usepackage{booktabs}
\usepackage{array}

\usepackage{microtype}
\usepackage{amsmath}
\usepackage{inconsolata}
\usepackage{enumitem}
\usepackage{xcolor}
\usepackage{multirow}
\usepackage{booktabs}
\usepackage{colortbl}

\definecolor{mybgcolor}{HTML}{ea998f} 
\definecolor{mybgcolor2}{HTML}{c4d1f7}
\definecolor{mybgcolor3}{HTML}{ffecc1}
\definecolor{mybgcolor4}{HTML}{aec5a8}

\newcommand{\model}{TSE\xspace}

\title{Blind Spot Navigation in Large Language Model Reasoning\\ with Thought Space Explorer}

\author{
    Jinghan Zhang\textsuperscript{\rm 1},
    Fengran Mo\textsuperscript{\rm 2},
    Tharindu Cyril Weerasooriya\textsuperscript{\rm 3},\\
    \textbf{Xinyue Ye}\textsuperscript{\rm 4},
    \textbf{Dongjie Wang}\textsuperscript{\rm 5},
    \textbf{Yanjie Fu}\textsuperscript{\rm 6},
    \textbf{Kunpeng Liu}\textsuperscript{\rm 1}\thanks{Corresponding author.}\\
    \textsuperscript{\rm 1}Clemson University,
    \textsuperscript{\rm 2}Université de Montréal,
    \textsuperscript{\rm 3}Center for Advanced AI, Accenture,\\
    \textsuperscript{\rm 4}The University of Alabama,
    \textsuperscript{\rm 5}University of Kansas,
    \textsuperscript{\rm 6}Arizona State University\\
  \texttt{\{jinghaz,kunpenl\}@clemson.edu} \\
}
\begin{document}
\maketitle
\begin{abstract}
Large language models have shown strong reasoning capabilities through chain-structured methods such as Chain-of-Thought. Recent studies optimize thought structures by generating parallel or tree-like structures, switching between long and short reasoning modes, or aligning reasoning steps with task performance. However, these approaches mainly rely on previously generated logical directions of the chains, which ignore the unexplored regions of the solution space. 
Such a phenomenon is defined as 
\emph{blind spots}, which limit the diversity and effectiveness of the reasoning process. 
To this end, we propose the ``Thought Space Explorer'' (TSE), a framework for navigating and expanding thought structures to overcome blind spots in LLM reasoning. 
Our TSE first identifies key nodes with high impact, then generates new nodes by integrating information from multiple chains. Finally, it extends new branches through connection strategies. 
We conduct a series of experiments on math and QA benchmarks. Compared with existing baseline methods, \model improves the accuracy of both the final answer and intermediate reasoning steps, while maintaining a better effectiveness-efficiency trade-off for practical deployment.
%final answer accuracy, enhances reasoning path accuracy and diversity while maintaining practical token-efficiency trade-off.
\end{abstract}

\section{Introduction}
\begin{figure*}
    \centering
    % \vspace{-8mm}
    \includegraphics[width=0.95\textwidth]{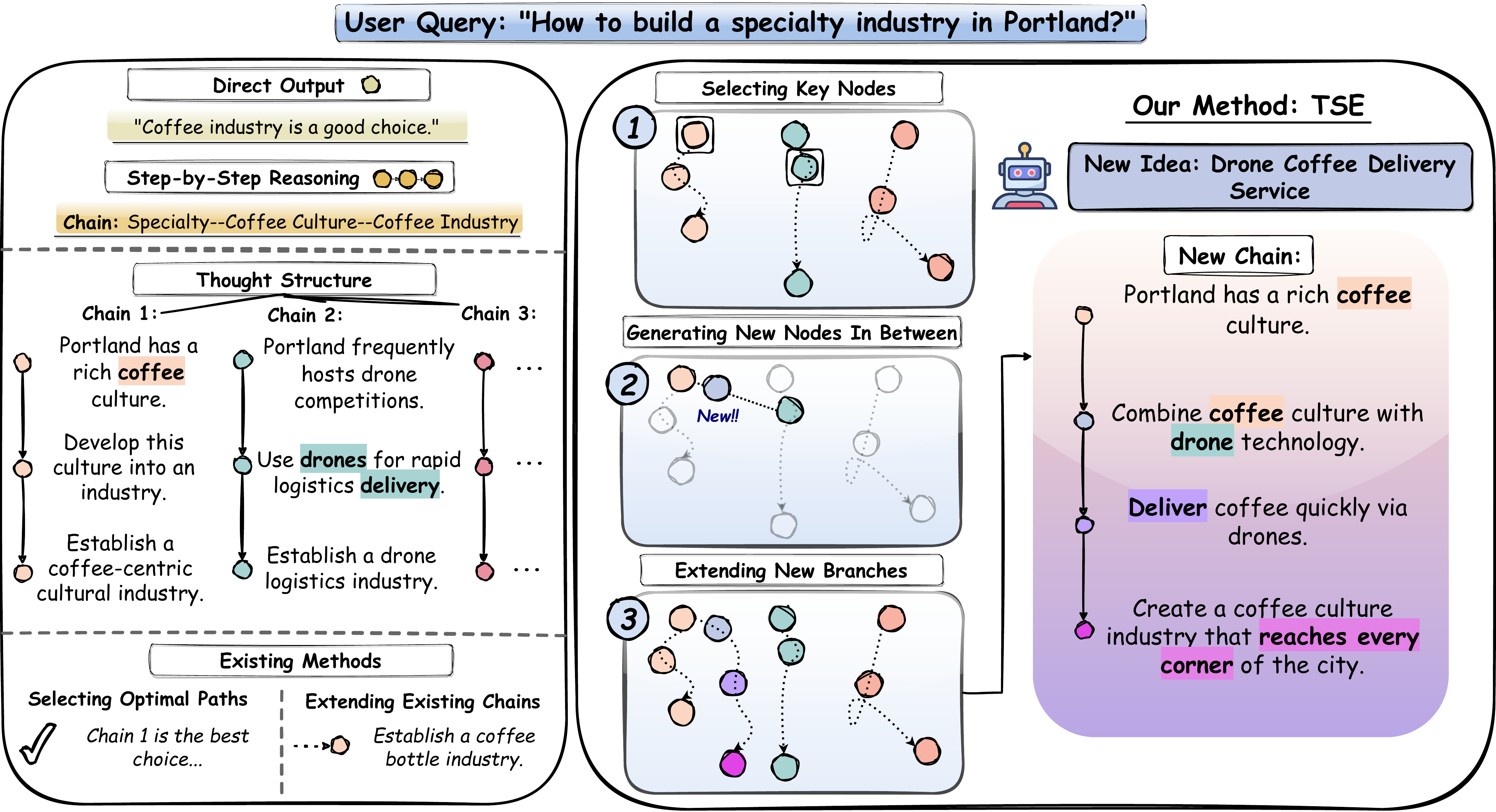}
    % \vspace{-2mm}
    \caption{Thought structure optimization through \model. On the left side, we showcase traditional thought structures and optimization methods, where the LLMs' generation may limited by its thought pattern. On the right side, we show how \model expands thought structure through a three-step generation of branches. \model guides LLMs to explore the blind spots between previous thought paths.}
    \label{fig:intro}
    % \vspace{-6mm}
\end{figure*}

Recent advances in large language models (LLMs) have shown great potential in solving complex tasks with reasoning capabilities~\citep{huang2022towards,patterson2022carbon,achiam2023gpt,mao2023largelanguagemodelsknow,dutta2025annotator} by guiding the LLMs to logically solve the complex task step-by-step. 
A common practice is to design the Chain-of-Thought (CoT)~\citep{kojima2022large,yang2025qwen3} to boost reasoning capabilities by evolving the thinking from a direct output to a a chain of intermediate reasoning steps.
Existing studies~\citep{wang2022self,yao2024tree,zhang2024diagram,besta2024graph,pandita2025prorefine} attempt to develop various thought structures with multiple chains or branches of thought on top of CoT to arouse the reasoning ability of LLMs.
Compared with direct output and CoT, the core advantage of thought structures enables models to explore the solution space of a task from local to global~\citep{hao2023reasoning}. For example, as presented in Figure~\ref{fig:intro}, thought structures may initiate exploration from two distinct points \textit{``specialty''} and \textit{``industry''}. Such exploration allows LLMs to generate diverse paths to solutions and thus enhances the model's reasoning capacity. 
Moreover, the diverse structures can enable models to perform forward and backward evaluations within the explored thought space toward the optimal solution, i.e., a more effective reasoning thought path. %These multi-dimensional thought structures not only perform high-quality outputs but also better leverage LLMs' potential for information mining and exploration.

% Existing works on optimizing thought structures primarily focus on extending current frameworks or selecting optimal paths. 
% Logic-of-Thought~\citep{}, which injects logic into context for a more comprehensive reasoning process. 

%Preliminary works on developing and optimizing thought structures includes 
A series of studies are conducted to optimize thought structures with various aspects, including generating parallel thought~\citep{wang2022self}, constructing tree-structured reasoning topologies on top of CoT~\citep{yao2024tree}, and fine-tuning the LLMs with direct preference optimization (DPO) to align thought steps of CoT with task performance~\citep{zhang2024chain}, etc.
%Chain-of-Thought with Self-Consistency (CoT-SC)~\citep{wang2022self}, generates a parallel thought structure with multiple chains of thought.
%which generates a parallel thought structure with multiple chains of thought; 
%Tree-of-Thought (ToT)~\citep{yao2024tree} constructs a tree structure reasoning topology with CoT being the branches. Chain of Preference Optimization (CPO)~\citep{zhang2024chain}, which fine-tunes LLMs with direct preference optimization (DPO) to align thought steps of CoT with ToT. 
%Aim for further improvement, 
The key idea of these studies is to compare multiple responses or extend existing chains (e.g., \textit{``Coffee over Drone''} or \textit{``Coffee industry $\rightarrow$ Coffee bottle industry''} as shown in Figure~\ref{fig:intro}) to obtain a better thought chain.

%These works involve comparing multiple responses or extending existing chains (e.g. \textit{``Coffee over Drone''} or \textit{``Coffee industry $\rightarrow$ Coffee bottle industry''}). 
%Based on these methods, LLMs can enhance reasoning efficiency and some depth of thought on the generated thought structures. 
% However, these approaches lack exploration of unknown solution spaces, which might lead to consistent oversight of LLMs' cognitive blind spots
However, these approaches do not explore regions of the solution space that the model itself has never considered. We refer to such unexplored regions as the \emph{blind spots} of LLMs. These blind spots are the areas in the reasoning space that are systematically overlooked, because the model's generation is biased and always leads to the previously explored paths~\citep{zhang2024prototypical,sprague2024cotcotchainofthoughthelps,liu2025mindstepbystep}. 
Merely generating more chains does not enable LLMs to conceive of content previously unthought of. As described in Figure~\ref{fig:intro}, over-generated chains tend to repeat prior thought patterns, leading to two main issues: (1) the absence of feasible solutions. When such solutions lie in blind spot regions, repeatedly filtering or extending existing paths may converge to a local optimum (e.g., exploring only from a \textit{coffee} perspective); and (2) insufficient diversity—especially for open-ended questions, where existing methods have limited impact on exploring the thought space, and excessive extension or filtering might even reduce the diversity of responses (e.g., discarding feasible solutions or creating redundancy through repetitive thinking).

To address these issues, we propose the \textbf{T}hought \textbf{S}pace \textbf{E}xplorer (\model), a novel framework designed to expand and optimize thought structures. 
The \model starts from thought paths already explored and guides the model to explore hidden solution spaces because the existing thought structures often already contain feasible solutions or crucial information pointing towards such solutions. %(e.g., connecting \textit{``Portland''} with \textit{``coffee culture''}).
% Extracting this information can serve as anchors for further reflection by the LLMs, avoiding blind exploration. Similar to the human process of innovation, we aim to use these ``spark lights'' from different thoughts as conditional information to break the model's habitual thinking patterns.
To enhance efficiency and precision, further exploration of the model starts from thought nodes within explored solutions, which ensures that the reasoning process is not a blind exploration but a deeper inquiry based on verified insights. %and thus exploring and leveraging hidden solution spaces more effectively.
% Specifically, to identify key points of information from existing thoughts based on ``which details have the greatest impact on the outcome'', and then attempt to connect multiple details to generate new thoughts. 

To identify key points of information from existing thoughts, as shown in Figure~\ref{fig:intro}, we first quantify each thought node's contribution to the conclusion during the model's reasoning process to select key nodes (e.g., in Chain 2, the details about \textit{``drones and delivery''} to serve as key information leading toward \textit{``logistics industry''}). 
% Considering the visibility of parameters in LLMs, we select key nodes from two perspectives: relative gradients and semantic relationships. 
We adopt relative gradients as an importance metric to select key nodes. Based on these key nodes, the model then generates new thought nodes and proceeds with deeper reasoning in new directions from ``original nodes'' to ``new nodes'', facilitating exploration of the solution space through the thought structure. Finally, we perform collaborative reasoning across the entire thought structure to generate the output. Considering the visibility of parameters in LLMs, we reformulate the key steps of this method using LLMs' semantic and evaluation capabilities for black-box or gradient-invisible models.

% Concretely, to identify key points of information from existing thoughts, we first quantify the contribution of each thought node to the conclusion during the model's reasoning process to select key thought nodes (e.g., in Chain 2, the details about \textit{``drones and delivery''} serve as key information leading toward \textit{``logistics industry''}). 
% Considering the visibility of parameters in LLMs, we select key nodes from two perspectives, relative gradients and semantic relationships. 
% Further, based on these key nodes, the model generates new thought nodes and proceeds with deeper reasoning in new directions from ``original nodes'' to ``new nodes''
% %as ``original node $\rightarrow$ new nodes'', 
% and facilitates the exploration of the solution space by the thought structure. Finally, depending on whether the reasoning tasks require a singular or comprehensive conclusion, we proceed with collaborative reasoning across the entire thought structure to generate the output. 
We evaluate the effectiveness of \model on four reasoning benchmarks on Qwen3 series models~\cite{yang2025qwen3} and the results show that \model significantly improves the performance of thought structures compared with existing methods. We further discuss the impact of \model on path accuracy beyond final output accuracy, the diversity enhancement and the token cost trade-off.

%and compare the impact of strategy selection in each stage. %In summary, we study the thought structure optimization problem and make the following technical contributions:

Our contributions include: 
\begin{itemize}
    \item We propose the \model reasoning framework which expands thought structures for exploring solution spaces to alleviate the impact of blind spots for LLM reasoning.
    \item We investigate various strategies to prioritize and refine the thought structure by identifying the importance of nodes in the thought structure.
    \item Experimental results on three specific reasoning tasks indicate the effectiveness of our \model compared with the existing reasoning methods without exploring thought structure.
\end{itemize}
\section{Related Work}
\subsection{LLM Reasoning Structures}
%For reasoning tasks, the most straightforward method is to prompt an LLM to generate a conclusion through one-time thinking. However, this direct approach often yields poor performance, as the LLM may overlook essential intermediate steps and generate incoherent logic or incorrect conclusions~\citep{chu2023survey}. 
The most straightforward method to address reasoning tasks is to generate a conclusion through one-step thinking. However, the LLM might overlook essential intermediate steps and generate incoherent logic or incorrect conclusions~\citep{chu2023survey}.
The advent of CoT~\citep{wei2022chain,wang2024chain} optimizes the reasoning step by connecting distinct thoughts into a coherent sequence~\citep{li2024chain}.
%has significantly transformed LLM reasoning by turning the isolated, point-based process into a linear, phased reasoning path that connects distinct thoughts into a coherent sequence~\citep{li2024chain}. 
Although CoT can improve transparency and coherence, its singular structure limits its capability to handle more complex logical relationships~\citep{jin2024self,sprague2024cotcotchainofthoughthelps}. 
To this end, some studies develop structured reasoning methods, such as self-consistent CoT and Tree-of Thought.~\citep{wang2022self,zhang2024diagram,yao2024tree,liu2024logic,mo2024tree,zhang2024ratt,zhang2024prototypical}. 
These sophisticated thought structures enhance the consistency and systematic nature of reasoning by expanding the ability of the model to manage diverse logical relationships~\citep{xia2024beyond,stechly2024chain}. 
Thought structures offer distinct advantages by maintaining coherence and depth while increasing the diversity and flexibility of reasoning paths~\citep{liang2024internal}. However, the reasoning chains within these structures are highly repetitive and thus reduce generation and selection efficiency.

% (Add one sentence to their disadvantage to motivate our thought structure expansion/Thought Structure Optimizations.)
%Furthermore, these structures enable LLMs to better adapt to and accurately solve a wide range of problems with varying complexities~\citep{liang2024internal}.

\subsection{Thought Structure Optimizations}
To further enhance the capabilities of thought structures, recent researches focus on two main optimization strategies. The first is selecting optimal paths within the structures~\citep{feng2023alphazero,long2023large,hao2023reasoning,shinn2023reflexion,jung2022maieutic}. By choosing optimal paths, the model filters the irrelevant and low-quality branches and globally searches for correct or optimal solutions, thus enhancing reasoning efficiency and accuracy. The second is expanding reasoning depth and breadth~\citep{zhu2022solving,besta2024graph,gao2024interpretable,zhang2024ratt,zhang2024chain,hou2024timetom,zhang2025entropy}. By deepening and widening the thought structure, the model can explore a broader array of possibilities and perspectives, thus improving its understanding and capability to solve complex issues~\cite{chen2025towards,pan2025survey,wang2025multimodal}. 
However, these methods might be limited to previously explored spaces and directions, thus failing to adequately investigate the blind spots within the thought space of the model.
%However, these methods are often limited to previously explored spaces and directions and fail to adequately investigate the blind spots within the model's thought space. Inspired by these two strategies, 
Different from them, we aim to expand the depth and breadth of thought structures by actively exploring the cognitive blind spots of the model in thought space. %while enhancing output quality through collaborative reasoning.
\section{Methodology}
To expand and optimize thought structures for effective exploration of reasoning spaces, we introduce \textbf{\model}, a self-expansion and exploration method that allows language models to proactively address deficiencies in reasoning processes and explore new reasoning directions with limited steps of generation.
We implement the \model through three stages: \textbf{(1) Key Node Selection and New Node Generation}, 
% in which we identify the most influential nodes and generate new nodes based on the crucial information they contain; 
which aims to select the nodes that are most influential to exploration directions based on the crucial information contained previously, then the model generates a new node to integrate the insights of two key nodes for new exploration directions;
\textbf{(2)} \textbf{New Node Connection and Chain Expansion}, 
% in which we systematically connect selected key nodes and new nodes, and expand them into new branches to explore new reasoning directions; 
to connect and expand the reasoning paths, and the new paths explore potential new directions of solutions from the new node;
and \textbf{(3) Multi-branch Reasoning} to address deficiencies in the model's ability to synthesize and integrate diverse reasoning paths in different directions.

\subsection{Problem Formulation}

Given a specific reasoning task $\mathcal{Q}$, we apply a large language model (LLM) $\mathcal{L}$ to a structured reasoning process $\mathcal{S}$. %, which we represent as a thought structure $\mathcal{S}$.
This structure consists of multiple reasoning sentences as thought nodes, which are connected sequentially.
The set of all thought nodes is denoted as $\mathcal{T}$, where each node $T_{ij}$ represents the $j$-th reasoning step in the $i$-th thought chain. 
The thought structure $\mathcal{S}$ can be viewed as a directed graph consisting of vertices (thought nodes) $\mathbf{V}$ and edges (connections between consecutive nodes) $\mathbf{E}$. 
Formally, we define them as:
\begin{equation}
\begin{aligned}
\mathbf{V} &= \bigcup_{i=1}^N \bigcup_{j=1}^{K_i} \{T_{ij}\}, \quad K_i = |C_i|, \\
\mathbf{E} &= \bigcup_{i=1}^N \bigcup_{j=1}^{K_i-1} \{(T_{ij}, T_{i,j+1})\},
\end{aligned}
\end{equation}
% where $N$ and $K$ is the number of the thought chain $C$ and the number of the node $\mathbf{V}$ contained in a thought chain $C$, respectively.
where $N$ is the number of thought chains, and $K_i$ is the number of nodes contained in chain $C_i$. Then, the structure $\mathcal{S}$ is defined as:
\begin{equation}
\mathcal{S} = (\mathbf{V}, \mathbf{E}), \quad 
C_i = \langle T_{i1}, T_{i2}, \ldots, T_{iK_i} \rangle
\end{equation}
\raggedbottom
\begin{figure}
    \centering
    % \vspace{-5mm}
    \includegraphics[width=
\linewidth]{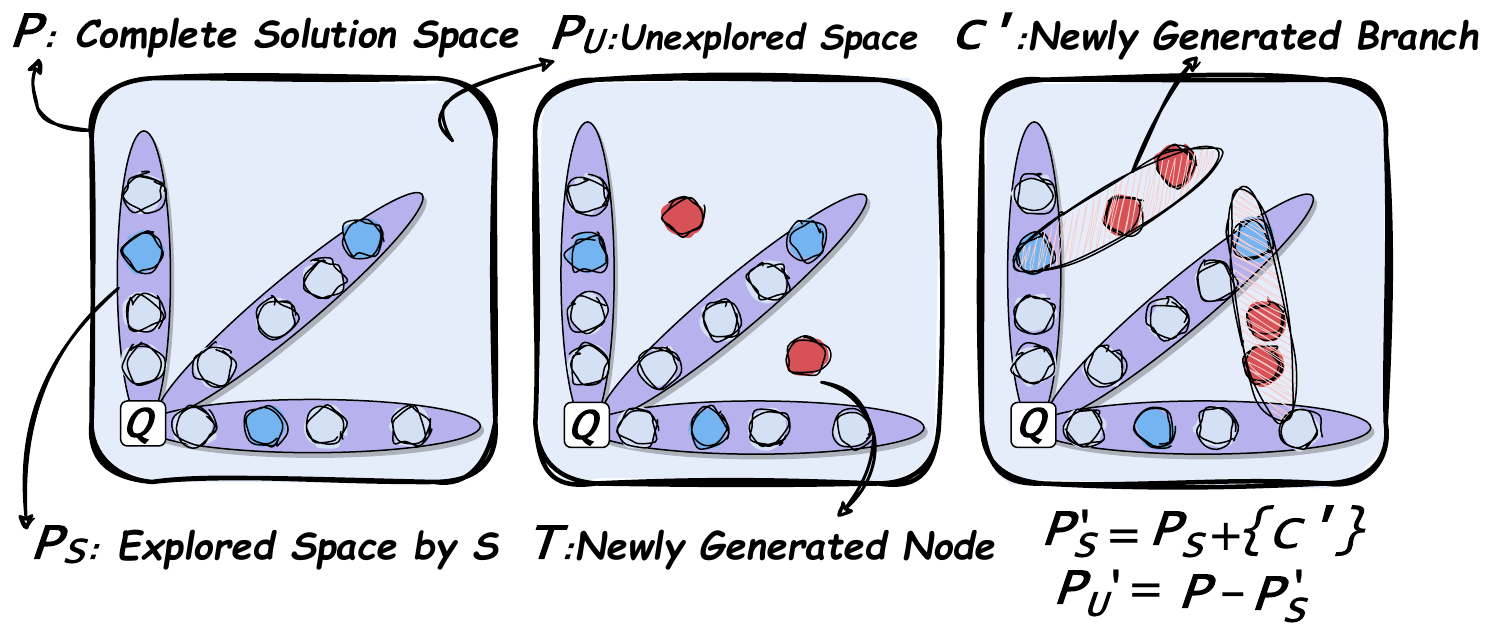}
    \caption{Illustration of solution space exploration via our \model method. By generating new branches of solutions, the explored space of solutions expands.}
    \vspace{-2mm}
    \label{fig:space}
    % \vspace{-4mm}
\end{figure}
For a specific task $\mathcal{Q}$, the complete reasoning solution space $\mathcal{P}$ encompasses all possible reasoning paths $C_i$ (thought chains) that can potentially solve $\mathcal{Q}$. 
As shown in Figure~\ref{fig:space}, the space that has been explored by the generated thought structure $\mathcal{S}$ is denoted as $\mathcal{P}_{S}$, and the remaining unexplored space is denoted as $\mathcal{P}_{U}$, with $\mathcal{P}_{S} \cup \mathcal{P}_{U} = \mathcal{P}$. 

\textbf{Our goal} is to actively expand the thought structure $\mathcal{S}$ by generating new reasoning branches $C'$ to explore previously untouched subspace $\mathcal{P_U}$. In this way, we increase the likelihood of discovering correct and novel solutions. Formally, we define the optimization objective as:
\begin{equation}
    \max_{\mathcal{S}'} J(\mathcal{S}', \mathcal{Q})
\end{equation}
where $J$ is the reasoning performance metrice and $\mathcal{S}'$ is the expanded thought structure.
% we maximize the chance of discovering effective and innovative solutions. The optimization goal is defined as $\max_{\mathcal{S}'} J(\mathcal{S}', \mathcal{Q})$, where $J$ is the reasoning performance metric and $\mathcal{S}'$ is the expanded thought structure.
% \begin{equation}
% \label{eq: optimization}
% \max_{\mathcal{S}'} J(\mathcal{S}', \mathcal{Q})
% \end{equation}

\subsection{Key Node Selection and New Node Generation}

We aim to select the most impactful nodes from the existing thought structure $\mathcal{S}$ for expansion. 
% Specifically, we expect the nodes to contain crucial information necessary for solving the given problem by achieving the following aspects: 
Intuitively, these nodes should contain crucial information for the task and satisfy two requirements:
(i) enabling effective exploration of promising regions in the solution space by initiating expansion from these key nodes, thus increasing the likelihood of discovering viable solutions; and 
(ii) reducing error propagation by conducting additional analysis and verification on these critical nodes, which often represent potential sources of mistakes.

% To accommodate the different availability of the backbone models, we propose two methods, gradient-based and self-prompting selection to identify the key nodes for those with accessible gradients and black-box ones, respectively.

\subsubsection{Gradient-based Selection} 
\begin{figure}[H]
\vspace{-2mm}
    \centering
    \includegraphics[width=\linewidth]{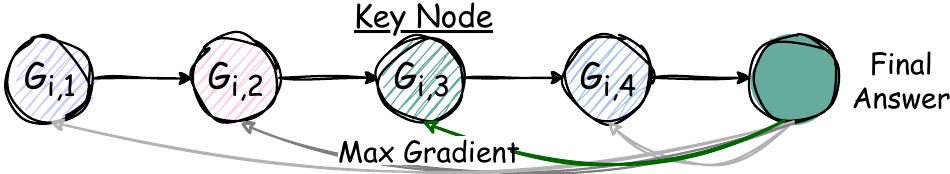}
    \caption{Gradient-based selection.}
    \label{fig:3keynode}
    \vspace{-3mm}
\end{figure}

% The gradient-based selection is applicable when the internal structure and the gradient information of the model $\mathcal{L}$ are accessible. The representation of each thought node $\mathcal{T}_{ij}$ in the generated thought chains is obtained via the model $\mathcal{L}$ as:
When the internal states and gradients of the model $\mathcal{L}$ are available, we access the representation of each thought node $T_{ij}$ from the hidden states of $\mathcal{L}$:
\begin{equation}
    \mathcal{L}: T_{ij} \mapsto \mathbf{v}_{ij} \in \mathbb{R}^d .
\end{equation}
% \raggedbottom
% Then, we aim to analyze the gradient importance of nodes $\mathcal{T}_{ij}$ relative to the conclusion node $\mathcal{T}_{iK_i}$ within a chain. 
The representation of the conclusion node $\mathbf{v}_{iK_i}$ is mapped to the output space as the model's prediction $\hat{y}_i$ as:
\begin{equation}
    \hat{y}_i = f(\mathbf{v}_{iK_i}),
\end{equation}
where $f(\cdot)$ denotes the mapping from the representation space to the output space, and $\hat{y}_i$ is typically a textual answer or decision for task $\mathcal{Q}$.

The self-information loss $L_i$ is a common practice to evaluate the model's confidence in its predictions~\citep{wang2021self}, where higher confidence corresponds to lower loss values.
Thus, we calculate the partial derivative of the loss $\mathbf{g}_{ij}$ with respect to each node's representation $\mathbf{v}_{ij}$ and the Euclidean norm of its gradient $G_{ij}$ to measure the importance of the nodes, as shown in Figure~\ref{fig:3keynode}.
Then, we apply a normalization to determine the relative importance $I_{ij}$ of each node for a consistent and comparative analysis of node importance across different chains within the thought structure $\mathcal{S}$ as:
% \begin{equation}
% L_i = -\log P(\hat{y}_i \mid \mathbf{v}_{iK_i}), \quad\mathbf{g}_{ij} = \frac{\partial L_i}{\partial \mathbf{v}_{ij}} 
% \end{equation}
% \begin{equation}
%     \quad G_{ij} = \|\mathbf{g}_{ij}\|_2, \quad I_{ij} = \frac{G_{ij}}{\sum_{k=1}^{K_i} G_{ik}}
% \end{equation}
\begin{equation}
\begin{aligned}
    L_i = -\log P(\hat{y}_i \mid \mathbf{v}_{iK_i}), 
    \\
    \mathbf{g}_{ij} = \frac{\partial L_i}{\partial \mathbf{v}_{ij}}, 
    \quad 
    G_{ij} = \|\mathbf{g}_{ij}\|_2 .
\end{aligned}
\end{equation}
% \raggedbottom
To enable comparison across nodes and chains, we normalize the gradient magnitudes within each chain $C_i$ as:
\begin{equation}
    I_{ij} = \frac{G_{ij}}{\sum_{k=1}^{K_i} G_{ik}} .
\end{equation}
% In this way, we regard nodes that have the highest impact on the model's predictions.
We regard nodes with larger $I_{ij}$ values as key nodes, as perturbations at these nodes typically have the greatest impact on the final prediction.
Each $T_{i}^{\text{key}}$ corresponds to the most influential node selected from chain $C_i$. 
These key nodes serve as the starting points for generating new reasoning branches in the subsequent expansion phase. The gradient-based selection use gradient magnitude as a joint indicator of information influence and uncertainty. It guides the model to explore from nodes with the highest potential gain. Meanwhile, it provides extra verification at nodes that are most likely to error with limited computing. In this way we make a balance between efficient exploration and stable control.
%We select the node in the chain with the highest relative gradient as the key node and obtain a set of key nodes $\mathcal{T}_{key} = \left\{ T_{i_{key}} \right\}_{i=1}^N$

\subsubsection{Generating New Nodes} 
With the selected key nodes, the next step is to use them as conditional information for generating new thought nodes. 
% Then we generate the new nodes based on the key nodes by integrating them into the thought structure and expanding them to form new branches. For each new node, we select two key nodes from $\mathcal{T}_{key}$, denoted as $\mathcal{T}_{i_{key}}$ and $\mathcal{T}_{l_{key}}$. With each pair of selected key nodes, the model generates a new thought node $\mathcal{T}^{1}_{il}$ as:
% \begin{equation}
% \mathcal{T}^{1}_{il} = \mathcal{L}(\mathcal{T}_{i_{key}}, \mathcal{T}_{l_{key}}), i, l \in [1, N] , i \neq l
% \end{equation}
% \raggedbottom
We generate the new nodes by combining two key nodes from the set $T^{\text{key}}$, denoted as $T_{i}^{\text{key}}$ and $T_{l}^{\text{key}}$. 
Given such a pair, the model generates a new candidate node $T^{1}_{il}$ as:
\begin{equation}
    T^{1}_{il} = \mathcal{L}(T_{i}^{\text{key}}, T_{l}^{\text{key}}), 
    \quad i,l \in [1,N], \, i \neq l .
\end{equation}

\subsection{Connection and Expansion}
\begin{figure}[H]
\vspace{-2mm}
    \centering
    \includegraphics[width=\linewidth]{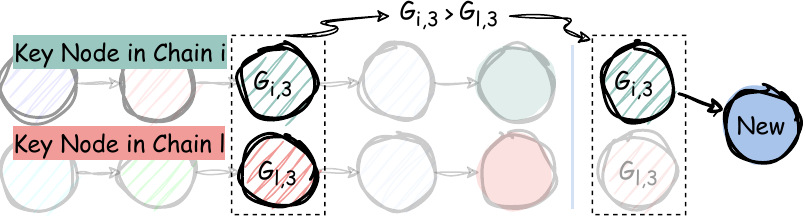}
    \caption{Gradient-based key node connection.}
    \label{fig:3connect}
    \vspace{-2mm}
\end{figure}
Then we integrate the new node into the thought structure. Since these key nodes are semantically closest to the new node, we choose between the two key nodes ($T_{i}^{\text{key}}$ or $T_{l}^{\text{key}}$) to decide which one can serve as the connection point for extending a new branch. 
% $\mathcal{T}_{i_{key}}$ or $\mathcal{T}_{l_{key}}$ to serve as the connection point $\mathcal{T}^{1}_{il}$ for extending the new thought branch, which is supposed to maintain logical coherence while exploring previously unknown regions of the thought space.
Therefore, we select the connection node as the key node that exhibits stronger semantic relevance to the newly generated node and contributes more significantly to reasoning. 
% Building on the previous stage, we propose two connection node selection methods, including the relative gradient method and the semantic relevance method.
% \subsubsection{Relative Gradient Selection} 
%Relative gradient selection is applied after gradient-based selection. For the normalized nodes $T_{i_{key}}$ and $T_{l_{key}}$, we select the node with the larger relative gradient as the connection point. This selection is made by comparing the importance indices $I_{i_{key}}$ and $I_{l_{key}}$:
The relative gradient selection chooses the connection node between $T_i^{\text{key}}$ and $T_l^{\text{key}}$ by comparing their importance indices $I(T_i^{\text{key}})$ and $I(T_l^{\text{key}})$. 
Formally, we first select the connection node and then initialize a new branch as:
\begin{equation}
\begin{aligned}
    T_c &= \arg\max_{T_{key} \in {T_i^{key}, T_l^{key}}} I(T_{key}),\\
    C' &= \langle T_c, T^1_{il} \rangle  .
\end{aligned}
\end{equation}
% \raggedbottom
where $C'$ denotes the new branch initiated from the key node with the higher importance index, as shown in Figure~\ref{fig:3connect}. Starting from the newly generated node $T^{1}_{il}$, the model $\mathcal{L}$ continues to generate subsequent steps conditioned on $T_c$. Since $T^{1}_{il}$ integrates information from two key nodes, the branch tends to explore novel reasoning directions that were not present in the original chains. 
The branch is extended until the target depth $K$ is reached. By default, $K$ is inherited from the chain containing $T_c$ (i.e., $K=K_i$ if $T_c=T_i^{\text{key}}$, otherwise $K=K_l$).

% Starting from $T^{1}_{il}$, the model $\mathcal{L}$ continues reasoning guided by the information from the new nodes, and since the new nodes introduce fresh perspectives of existing information, the new branch is likely to go and explore previously unconsidered directions with step-by-step reasoning.
% \begin{equation}
% \mathcal{L}(C') = \left\langle \arg\max_{T_{key} \in \{T_{i_{key}}, T_{l_{key}}\}} (I_{key}), T^{1}_{il}, T^{2}_{il}, \ldots, T^{K}_{il} \right\rangle
% \end{equation}
% \begin{equation}
% C' \rightarrow \mathcal{L}(C') = \langle C', T^{2}_{il}, \ldots, T^{K}_{il} \rangle,  \quad 
% K = \begin{cases} 
% K_i & \text{if } I_{i_{key}}, > I_{l_{key}} \\
% K_l & \text{otherwise}.
% \end{cases}
% \end{equation}

\raggedbottom

\subsection{Multi-branch Reasoning}
% By generating new nodes and branches, the model incorporates new reasoning paths into $\mathcal{S}$. The thought space:
Finally we reason and produce an output with both original and new branches.
Now we have a task $\mathcal{Q}$ and its complete but unseen solution space $\mathcal{P}$, the model $\mathcal{L}$ generates new thought branches on top of the original thought structure $\mathcal{S}$. 
During this process, each new branch $C'$ expands the explored subspace $\mathcal{P_S} \in \mathcal{P}$ by mining potential solutions based on the established structure as:
\begin{equation}
    \mathcal{P_S'} \leftarrow  \mathcal{P_S} \cup \{C'\}, \quad
    \mathcal{P_U'} \leftarrow \mathcal{P}-\mathcal{P_S'}.
\end{equation}
\raggedbottom
The refined structure $\mathcal{S'}$, compared to $\mathcal{S}$, explores a larger portion of the solution space with $|\mathcal{P_S'}| \geq |\mathcal{P_S}|$. 
Based on $\mathcal{S'}$, we can integrate both original and newly discovered reasoning paths to form a unified conclusion.
We consider all thought chains in the refined structure $\mathcal{S}'$ and use gradient information to recalculate and select the key nodes of each chain. For a key node $T_{ik}^{\text{key}}$, we assign a weight based on its relative contribution to the solution as:
\begin{equation}
    w_{ik}^{\text{key}} = \frac{\exp\!\left(-L_{ik}^{\text{key}}\right)}{\sum_{T_{im} \in T^{\text{key}}_i} \exp\!\left(-L_{im}^{\text{key}}\right)},
\end{equation}
\raggedbottom
where $L_{i_{key}}$ represents the self-information loss at node $\mathcal{T}_{i_{key}}$, which reflects the model’s confidence and potential error at that node. 
% The contribution of each key node, denoted as $\mathbf{v_{i_{key}}}$, is the value vector that quantifies the influence of the node on the overall reasoning process. 
The contribution of each node is represented as $v_{ik}^{\text{key}}$, which is obtained from its embedding through a linear projection.
This score quantifies how strongly the node supports the overall reasoning process, capturing factors such as semantic relevance to the task or inference correctness.  
% This vector can include various factors such as the relevance of the node's content to the task or the accuracy of its inference. 
%For the reasoning task $\mathcal{Q}$, 
Then, we compute the collaborative reasoning score
% by summing the contributions of all key nodes across all chains as:
by aggregating the weighted contributions of all key nodes across all chains for the given reasoning task $\mathcal{Q}$ as:
\begin{equation}
    C(\mathcal{Q}) = \sum_{i=1}^{N} \sum_{T_{ik} \in T^{\text{key}}_i} w_{ik}^{\text{key}} \cdot v_{ik}^{\text{key}} .
\end{equation}
\raggedbottom
% This process integrates the weighted contributions for a reasoning score that involves the individual node's direct impact and its significance within the context of the entire chain. The final decision $D$ for the task $\mathcal{Q}$ is selected based on the highest collaborative reasoning score via 
% \begin{equation} 
%     D = \arg\max_{q \in \mathcal{Q}} C(q)
% \end{equation}
So far, the decision $D$ for task $\mathcal{Q}$ is selected as the candidate with the highest collaborative reasoning score as:
\begin{equation}
    D = \arg\max_{q \in \mathcal{Q}} C(q).
\end{equation}
\raggedbottom
% blackbox situation

\subsection{Gradient-agnostic Blind Spot Navigation}
In gradient-available models, \model leverages internal gradients to identify and expand key reasoning nodes. However, in many practical scenarios, the proprietary LLMs are only accessible through APIs, where internal states and gradients cannot be observed. To adopt \model in such cases, we reformulate the key steps of exploration using the semantic and self-evaluation capabilities of LLMs. This module consists of three components: (i) Self-prompting Selection of key nodes, (ii) Semantic Relevance–based Connection Selection, and (iii) LM-as-a-judge for final reasoning evaluation.

\subsubsection{Self-prompting Selection} 
For a black-box model, we leverage LLM's capabilities to rank nodes within each path by their semantic and logical relevance to the task. 
Concretely, for a thought chain $C_i$, the model $\mathcal{L}$ evaluates each node $T_{ij}$ under the task context $\mathcal{Q}$ and assigns an importance score. 
The most critical node with the highest score is then selected as the key node $T_{i}^{\text{key}}$:
% analyze and prioritize key nodes in the thought chains. Although the inner mechanisms of the model are opaque, we infer critical areas within the network's structure by constructing specific prompts based on semantic and logical relationships. Specifically, model $\mathcal{L}$ ranks the importance of nodes within chain $C_i$ for the key node as:
\begin{equation}
    T_{i}^{\text{key}} = \arg\max_{T_{ij} \in C_i} \text{Rank}_{\mathcal{L}}(T_{ij} \mid \mathcal{Q}),
\end{equation}
where $\text{Rank}_{\mathcal{L}}(\cdot)$ denotes the ranking mechanism induced by carefully designed prompts that ask the model to compare nodes according to their semantic contribution and logical necessity.

\subsubsection{Semantic Relevance–based Connection Selection} 
% Semantic relevance selection is applicable after both key node selection methods. The connection node is selected by the model $\mathcal{L}$ based on semantic relationships between nodes. Then the new branch $C'$ continues to extend recursively until a specified depth is reached, with $\mathcal{L}$ generating subsequent nodes based on the strongest semantic relationship as:
After obtaining the key nodes, we need to decide which one should serve as the connection point when generating new branches. 
To achieve so, we rely on semantic relevance to enable the model $\mathcal{L}$ to compare the newly generated node $T^{1}_{il}$ with its candidate parent nodes $T_i^{\text{key}}$ and $T_l^{\text{key}}$, and select the node with stronger semantic alignment as the connection point $T_{\text{selected}}$ as:
\begin{equation}
    C' = \langle T_{\text{selected}},\, T^{1}_{il},\, T^{2}_{il},\, \ldots,\, T^{K}_{il} \rangle,
\end{equation}
where $K$ denotes the maximum depth of expansion. 
This ensures that the extended branch remains logically coherent while exploring novel directions.

% where $\mathcal{T}_{\text{selected}}$ is the key node chosen by $\mathcal{L}$, and $K$ denotes the ending depth of the thought chain.

\subsubsection{LLM-as-a-Judge} 
% This strategy is suitable for reasoning tasks that require detailed interpretation and judgment, applicable to both gradient-available models and black-box models. Here, the model $\mathcal{L}$ acts as a judge to score each thought chain's output based on its assessment of reasoning coherence, prediction confidence, or relevance to the task. Based on the evaluation, the model employs a voting mechanism to determine the final output:
Finally, for reasoning tasks that require interpretability or involve multiple competing branches, we employ the LLM itself as the judge to evaluate and aggregate the reasoning paths.
In this stage, model $\mathcal{L}$ assigns a score to each candidate chain $C_q$ based on specific factors, including soundness, coherence, innovation, and clarity. Then, the judgment scores are produced via an evaluation template as:
\begin{equation}
    \text{Score}_q = {\mathcal{L}}(C_q | \text{Eval Template}).
\end{equation}
The final decision $D$ for task $\mathcal{Q}$ is chosen through a voting or max-score mechanism:
\begin{equation}
    D = \arg\max_{q \in \mathcal{Q}} \text{Score}_q .
\end{equation}
In this way, these steps allow exploration and validation without access to gradients, and guide the model to more diverse and reliable reasoning paths.

\section{Experiments}
We conduct a series of experiments to evaluate the reasoning performance of \model. We first compare it with state-of-the-art baseline methods on several widely used benchmarks to test its overall accuracy. %which is the most important metric for reasoning. 
Then we investigate how \model enhances the quality of reasoning paths beyond the final answers to the questions, which includes validating the path accuracy (the effectiveness of the reasoning exploration) and the path diversity (whether the exploration leads the model search to broad paths for the final answer). Finally, we discuss the cost-accuracy trade-off of these methods. For more discuss and comparison of black box model and non-gradient exploration, please refer to \ref{sec:appendix}.
\subsection{Experimental Setup}
\paragraph{Settings.} We evaluate the pass@1 rate on the latest \texttt{Qwen3-4B/8B}~\cite{yang2025qwen3} models with sampling temperature fixed at 0.7. Unless otherwise specified, we generate five parallel thought chains with a maximum depth of 5 for each question and use this structure as the basis for \model. All experiments are conducted on 4 H200 GPUs.
\paragraph{Evaluation Datasets.} For evaluation, we select four widely used math and science benchmarks with different level: \textbf{GSM8K}~\cite{cobbe2021training}, a large collection of grade-school math word problems testing multi-step arithmetic reasoning. \textbf{AIME24} and \textbf{AIME25}~\cite{mathai_aime24,mathai_aime25}, each containing 30 competition-style math problems covering arithmetic, algebra, and geometry from the American Invitational Mathematics Examination; and \textbf{GPQA-D}iamond~\cite{rein2024gpqa}, a curated subset of GPQA, which contain 198 PhD-level science questions authored by domain experts in physics, chemistry, and biology. 
\paragraph{Baselines.}
We compare \model method with (1) \textbf{Direct} output without thinking mode, (2) \textbf{Think}~\cite{yang2025qwen3}: output with Qwen3's long CoT thinking mode, (3) \textbf{ToT})~\cite{TreeOfThoughts}: a reasoning method with tree-structure for thoughts, (4) \textbf{RATT}~\cite{zhang2024ratt,semnani2023wikichat}: a tree-structure reasoning method with RAG with WikiPedia as external knowledge base, and (5) \textbf{Self}-Route~\cite{he2025self}: an automated reasoning path mode switch method.
For ToT and RATT, we set the number of nodes with one generation ($N=3$). For Self-Route, the router uses MATH-500~\cite{lightman2023let} and GPQA non-diamond subset for training.
\subsection{Experiment Results}

\paragraph{Overall Performance.}
Our main results for accuracy on four benchmarks are shown in Table~\ref{tab:acc-improve-colored}. We find that our TSE method consistently demonstrates superior effectiveness compared to other reasoning approaches. For math tasks in GSM8K and AIME25, TSE achieves the highest accuracy for both Qwen3-4B and Qwen3-8B. In AIME24, \model remains the highest in the 4B model setting and on par with Self-Route by underperforming only one question in 8B model setting.
%in 8B, \model loses only one more question compared to Self-Route. 
For the QA task, \model remains the second-best competitive performance, since the best approach RATT uses retrieval and can access external scientific knowledge to answer complex questions. On average, \model reaches 59.2\% and 51.6\% accuracy improvement compared to the non-thinking outputs in terms of Qwen3-4B and Qwen3-8B, respectively. These results indicate that \model is effective in solving tasks of various difficulties and explores the correct reasoning directions sophisticatedly.
\begin{table*}
\centering
\caption{Accuracy (\%) and relative improvement over Direct baseline across four benchmarks. Best accuracy is shaded, second-best is \underline{underlined}.}
\label{tab:acc-improve-colored}
% \small
\begin{tabular}{llcccccccccc}
\hline
\toprule

\textbf{Model} & \textbf{Method} & \multicolumn{2}{c}{\textbf{GSM8K}} & \multicolumn{2}{c}{\textbf{AIME24}} & \multicolumn{2}{c}{\textbf{AIME25}} & \multicolumn{2}{c}{\textbf{GPQA-D}} & \multicolumn{2}{c}{\textbf{Avg.}} \\
& & Acc & ↑\% & Acc & ↑\% & Acc & ↑\% & Acc & ↑\% & Acc & ↑\% \\
\midrule
\multirow{6}{*}{Qwen3-4B}
& Direct      & 86.2 & --   & 20.0 & --   & 20.0 & --   & 36.4 & --   & 40.7 & --   \\
& Think       & 92.0 & 6.7  & 60.0 & 200.0 & \underline{48.9} & 144.5 & 45.0 & 23.6 & \underline{61.5} & 51.1 \\
& ToT         & 92.7 & 7.5  & \underline{63.3} & 216.5 & 40.0 & 100.0 & 46.0 & 26.4 & 60.5 & 48.8 \\
& RATT        & 92.7 & 7.5  & 56.7 & 183.5 & 46.7 & 133.5 & \cellcolor[gray]{0.9}\textbf{54.5} & 33.2 & 61.2 & 50.4 \\
& Self-Route  & \underline{93.1} & 8.0  & 56.7 & 183.5 & 46.7 & 133.5 & 43.0 & 18.1 & 59.9 & 47.2 \\
& TSE (Ours)        & \cellcolor[gray]{0.9}\textbf{94.0} & 9.0  & \cellcolor[gray]{0.9}\textbf{66.7} & 233.5 & \cellcolor[gray]{0.9}\textbf{50.0} & 150.0 & \underline{48.5} & 33.2 & \cellcolor[gray]{0.9}\textbf{64.8} & 59.2 \\
\midrule
\multirow{6}{*}{Qwen3-8B}
& Direct      & 88.5 & --   & 16.7 & --   & 23.3 & --   & 44.4 & --   & 43.2 & --   \\
& Think       & 93.4 & 8.1  & 46.7 & 179.6 & \underline{46.7} & 100.4 & 56.6 & 27.5 & 61.4 & 42.1 \\
& ToT         & 90.3 & 2.0  & 40.0 & 139.5 & 33.3 & 42.9 & 48.0 & 8.1 & 52.9 & 22.5 \\
& RATT        & 92.7 & 4.7  & 50.0 & 199.4 & 36.7 & 57.5 & \cellcolor[gray]{0.9}\textbf{59.1} & 20.5 & 58.2 & 34.7 \\
& Self-Route  & \cellcolor[gray]{0.9}\textbf{96.0} & 8.5  & \cellcolor[gray]{0.9}\textbf{63.3} & 278.4 & 43.3 & 85.8 & 55.1 & 24.1 & \underline{64.4} & 49.1 \\
& TSE (Ours)         & \underline{95.7} & 5.5  & \underline{60.0} & 239.5 & \cellcolor[gray]{0.9}\textbf{53.3} & 128.8 & \underline{58.6} & 32.0 & \cellcolor[gray]{0.9}\textbf{65.5} & 51.6 \\
\hline
\bottomrule
\end{tabular}
\end{table*}

\paragraph{Path Accuracy.} We then investigate how \model improves the accuracy of reasoning paths. Beyond providing correct final outputs, the ideal reasoning process should also provide logically clear and accurate intermediate steps for users to verify the trustworthiness. Thus, we output the reasoning chains generated by the Qwen3-8B model on GSM8K and evaluate their correctness by \texttt{GPT-4o}~\cite{hurst2024gpt}. The principle is that the GSM8K contains relatively simple math word problems with deterministic answers and requires pure arithmetic reasoning, and GPT-4o can effectively validate each step's correctness.
\begin{figure}[t]
    \centering
    \includegraphics[width=0.9\linewidth]{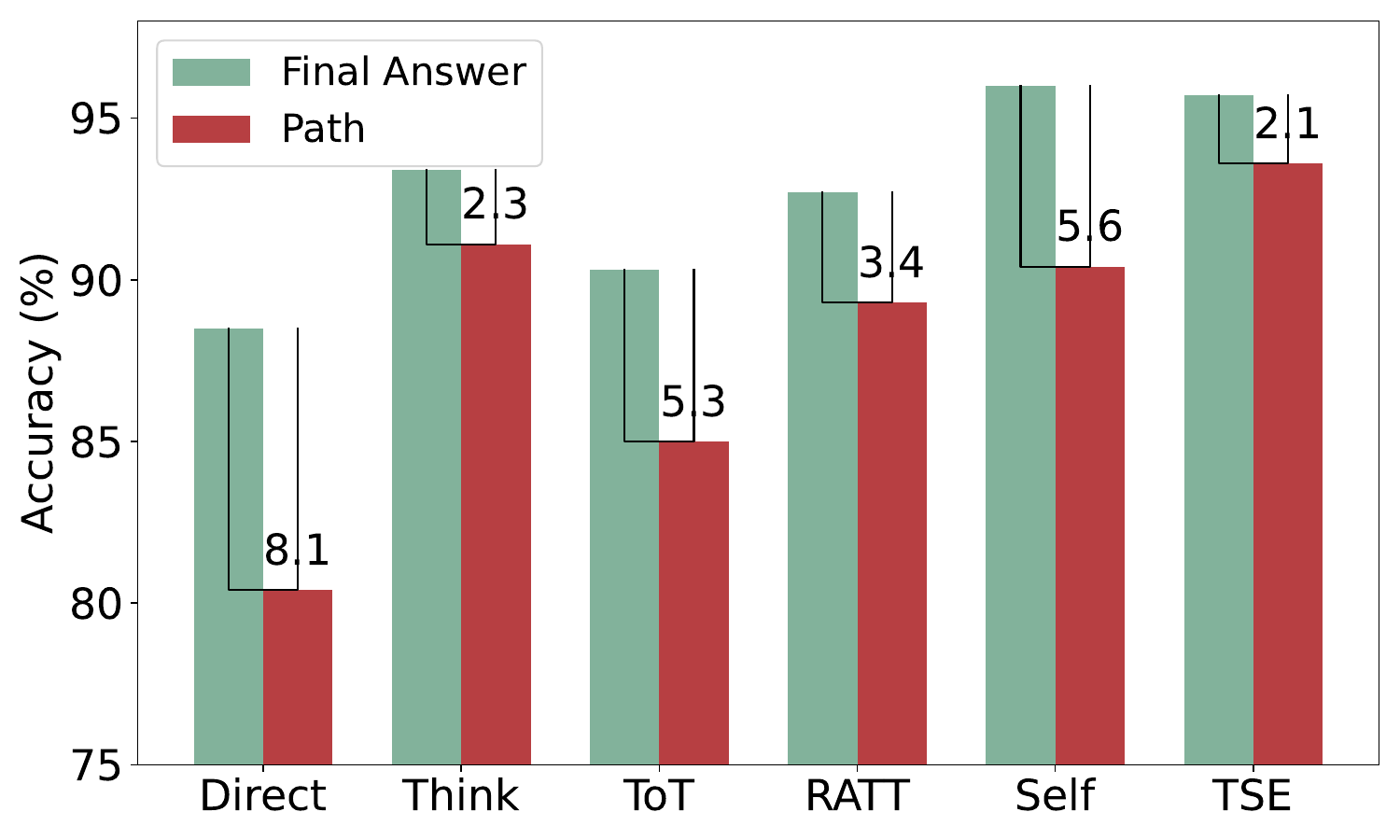}
    \caption{Path accuracy compared with final answer accuracy on GSM8K with Qwen3-8B.}
    \label{fig:pathacc}
\end{figure}

As shown in Figure~\ref{fig:pathacc}, we observe that in some cases the final answer is correct while the reasoning path contains errors. 
This phenomenon further highlights the importance of monitoring and correcting the process rather than just the results. The results indicate that \model not only achieves the highest final answer accuracy but also has the best path accuracy, which demonstrates its superiority in generating reasoning paths that are logically more consistent and verifiable. 
\paragraph{Diversity Measurement.} We further validate how \model's exploration contributes to the diversity of reasoning. In our experiments, we use Jaccard Similarity~\cite{niwattanakul2013using} to measure the diversity of reasoning trajectories produced by different methods. We segment each reasoning chain into a set of steps by delimiters or step indices. For a given question, we randomly select three reasoning chains that are both correct in answer and path as reference chains, and compute the Jaccard Similarity with all other chains. The Jaccard similarity is defined as
\[
J(C_i, C_l) = \frac{|C_i \cap C_l|}{|C_i \cup C_l|},
\]
where the intersection $\cap$ represents the shared steps and the union $\cup$ represents the total distinct steps. 
A higher Jaccard Similarity indicates a larger overlap between generated chains and thus results in lower diversity among generations. %while a low value indicates greater differences and higher diversity. 
For multiple reasoning chains, we compute the Jaccard Similarity for all chain pairs and take the average. We adopt the complement $1-\bar{J}$ as the overall diversity metric for reasoning paths.

\begin{figure}[htbp]
    \centering
    \includegraphics[width=\linewidth]{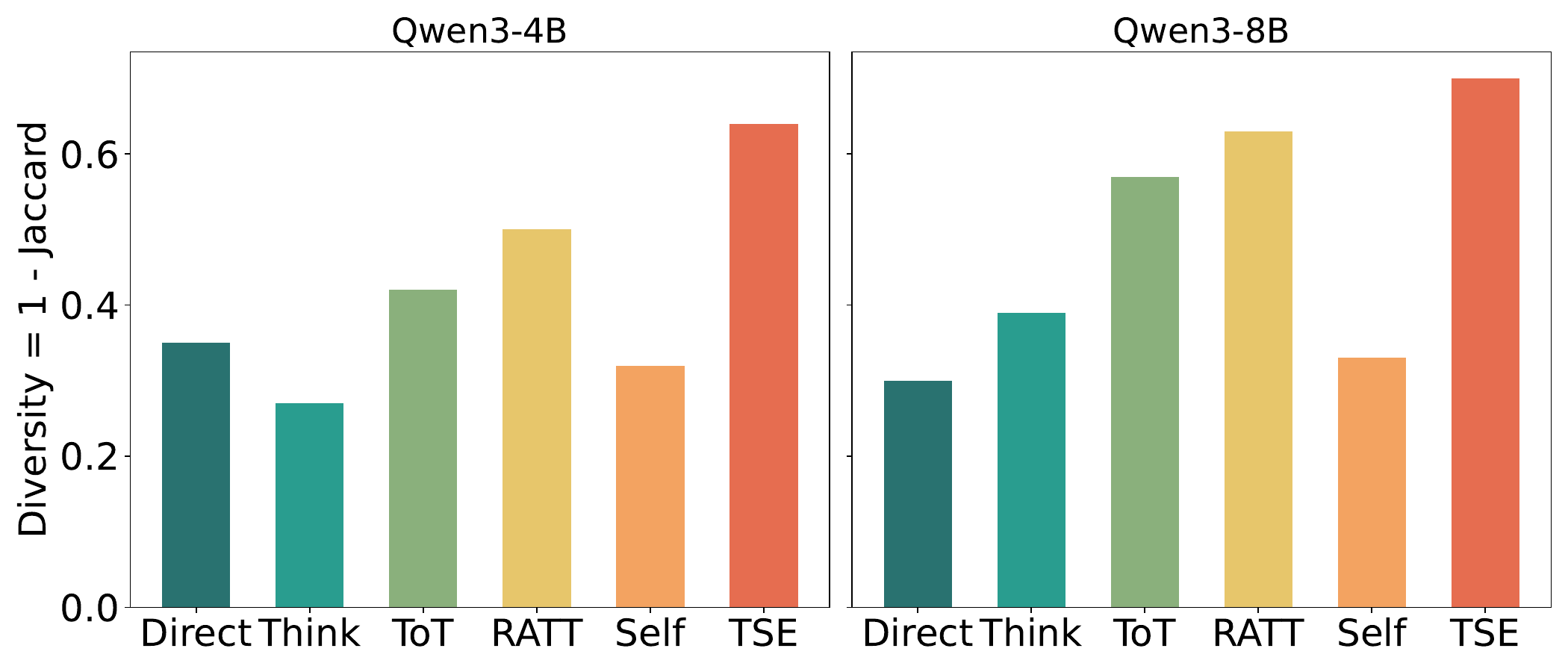}
    \caption{Reasoning diversity across different methods on GPQA-D with Qwen3-4B/8B models.}
    \label{fig:diversity}
\end{figure}

As shown in Figure~\ref{fig:diversity}, for both Qwen3-4B and Qwen3-8B models on the GPQA-D dataset, \model has the highest diversity, surpassing ToT and RATT that contain complex tree structures and naturally have wider exploration fields. Meanwhile, the Self-Route method, although it has competitive output accuracy, its reasoning paths are less diverse and have low path accuracy. Such observations indicate that \model explores genuinely new reasoning directions rather than simply expanding existing ones. More importantly, \model translates this exploration into higher-quality reasoning trajectories, thus enhancing both diversity and correctness that other methods might fail to realize.

\begin{figure}[H]
    \centering
    \includegraphics[width=0.9\linewidth]{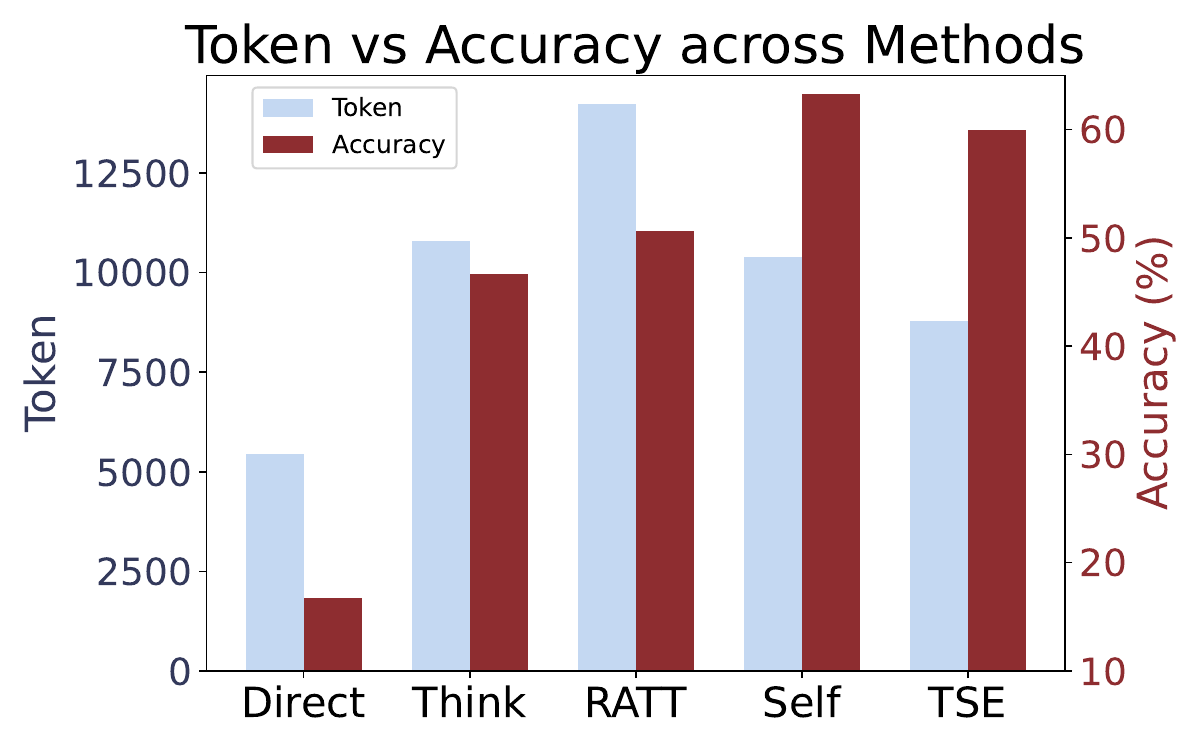}
    \caption{Token usage against accuracy on AIME24 with Qwen3-8B.}
    \label{fig:token}
\end{figure}

\paragraph{Token Usage.}
As \model requires additional exploration during content generation, we further analyze its computational cost. As shown in Figure~\ref{fig:token}, the direct output baseline without generating reasoning tokens has the lowest cost but substantially lower accuracy than all other methods. 
The Think method, with more than twice the tokens, achieves higher accuracy but still lower than the RATT and Self-Route methods. 
For ToT, as the number of generated nodes grows exponentially, it has an extremely high token consumption than others, which is not comparable, so we omit it from the figure. The RATT has the highest token usage as it contains a retrieval process, which may not help in arithmetic for math tasks, but can significantly improve QA task performance. Self-Route achieves better accuracy with fewer tokens compared to Think. For \model, it has competitive accuracy and exhibits the best token-accuracy trade-off, which demonstrates \model's better effectiveness-efficiency trade-off and its practical value for deployment under limited resources.

\section{Conclusion}
In this study, we introduce \model, a novel approach to enhance the reasoning structures of LLMs. \model generates new thought branches based on existing thought paths to explore previously overlooked solutions.
The generated new reasoning nodes and chains are incorporated into thought structures to explore diverse reasoning directions in terms of a reasoning task.
%In \model's framework, we systematically generated new reasoning nodes and chains, incorporating them into thought structures to explore diverse reasoning directions to the task. 
Our experiments across multiple reasoning datasets demonstrate the effectiveness of the \model. 
The detailed analysis reveals the utilization of each component in \model during the integration of diverse thought processes.
%The results show both quantitative improvements in performance and qualitative enhancements during the integration of diverse thought processes.

\newpage
\section*{Limitations}

While TSE demonstrates consistent improvements in reasoning accuracy, path quality, and diversity, several limitations remain. 
First, the expansion process introduces additional generation cost: although TSE achieves a favorable token–accuracy trade-off compared with baselines, it still requires more tokens than direct decoding. 
Second, the semantic-based variant depends on prompting quality and the model’s intrinsic evaluation ability, which may yield unstable results across different black-box APIs. 
Third, our experiments are conducted on widely used benchmarks (GSM8K, AIME24/25, GPQA-Diamond); further validation on more diverse domains (e.g., legal or medical reasoning) is needed to assess generality. 
Finally, TSE focuses on static reasoning structures, whereas real-world tasks often require adaptive depth control or interactive feedback, which remain open directions. 

These limitations suggest promising avenues for future work, including developing lighter expansion strategies, designing more robust semantic evaluation protocols, and extending TSE to dynamic or domain-specific reasoning scenarios.

\bibliography{ref}

\clearpage

\appendix

\section*{Appendix}
\label{sec:appendix}

\section{Experiments Comparing Black-box and White-box Models}

%We conduct three experiments on representative tasks to evaluate the effectiveness of our method. Here, we include three reasoning tasks that require mathematical, hierarchical, and comprehensive reasoning abilities, respectively. These tasks give an overview of \model's enhancement on thought structure, and we further compare the performance of each component and test their necessity.

\begin{table*}[htbp]
  \centering
  % \resizebox{0.88\textwidth}{!}{%
    % Make the text size smaller
    \begin{tabular}{c|c|c|c|c|c|c}
      \toprule
      \midrule
      \textbf{Task} & \textbf{Game of 24-GPT} & \multicolumn{5}{c}{\textbf{Creative Writing-GPT}} \\
      \hline
      & \textbf{Success Rate (\%)} & \textbf{Soundness} & \textbf{Innovation} & \textbf{Coherence} & \textbf{Expression} & \textbf{Overall} \\
      \hline
      CoT   & 13.30  & 5.26  & 5.24  & 5.03  & 4.95  & 5.12  \\
      \hline
      CoT-SC & 46.70  & 5.41  & 5.17  & 5.17  & 5.16  & 5.23  \\
      \hline
      ToT   & 52.70  & 5.90  & 5.36  & 5.40  & 5.43  & 5.52  \\
      \hline
      RATT  & \underline{41.30}  & \underline{6.02}  & \underline{6.22}  & \underline{5.69}  & \textbf{5.67}  & \underline{5.90}  \\
      \hline
      TSE   & \textbf{74.00}  & \textbf{6.19}  & \textbf{6.55}  & \textbf{5.74}  & \underline{5.64}  & \textbf{6.03}  \\
      \hline
      \midrule
      \textbf{Task} & \textbf{Game of 24-Llama} & \multicolumn{5}{c}{\textbf{Creative Writing-Llama}} \\
      \hline
      & \textbf{Success Rate (\%)} & \textbf{Soundness} & \textbf{Innovation} & \textbf{Coherence} & \textbf{Expression} & \textbf{Overall} \\
      \hline
      CoT   & 1.33  & 4.63  & 3.98  & 4.25  & 4.20  & 4.27  \\
      \hline
      CoT-SC & 5.33  & 4.84  & 4.35  & 4.51  & 4.46  & 4.54  \\
      \hline
      ToT   & 8.00  & 4.86  & 4.95  & 4.78  & 4.65  & 4.81  \\
      \hline
      RATT  & \underline{8.67}  & \textbf{5.10}  & \underline{4.97}  & \underline{4.89}  & \underline{4.97}  & \underline{4.98}  \\
      \hline
      TSE   & \textbf{18.75}  & \underline{5.03}  & \textbf{5.11}  & \textbf{5.05}  & \textbf{5.15}  & \textbf{5.09}  \\
      \hline
      \bottomrule
    \end{tabular}
  % }
  \caption{Overall performance on reasoning tasks based on GPT-4o-mini and Llama. \textbf{Bold} and \underline{underline} indicate the best and the second-best results.}
  \label{tab:overall}
  % \vspace{-2mm}
\end{table*}

% to address the following questions: 
% \begin{enumerate}
%     \item How does \model enhance the performance of existing thought structure in reasoning tasks?
%     \item How does the choice of method at each stage of the \model process affect the performance?
%     \item What is the impact of each component of the \model method?
% \end{enumerate}

% To answer these questions, we first evaluate the overall performance of \model compared to baseline methods on with different complexity. Then, we analyze the impact of key node selection, comparing gradient-based selection, model-based judgment, and random selection. Finally, we examine the contributions of the new chains generated by our method, highlighting how they differ from the original chains and how they enhance the coherence and innovation of the thought structure.

\subsection{Experimental Settings}
\par \textbf{Implementation Details and Baselines.} We conduct experiments on three reasoning tasks that require mathematical, hierarchical, and comprehensive reasoning abilities to evaluate the effectiveness of our \model method. We conduct these experiments utilizing the \texttt{GPT-4o-mini}~\citep{gpt4omini} and \texttt{Llama-3.1-8B-Instruct}~\citep{llama31}. 
Unless otherwise specified, we generate five parallel thought chains with a depth of 5 for each question and use this structure as the basis for new node generation. The temperature for all models is set to the default value of 0.7, with a maximum token limit of 50. All tasks are performed on an NVIDIA 4090 GPU. We apply our method to the simplest multi-chain thought structure and compare the results with several baseline methods, including CoT, Vanilla CoT-SC, ToT, and RATT. This comparison aims to illustrate how our method enhances the thought structure compared to existing approaches.

\noindent \textbf{Task Description.} We evaluate \model and the baseline methods on three task-specific reasoning datasets: (1) \textit{Game of 24} \cite{NEURIPS2023_271db992}, a mathematical challenge to use the four basic arithmetic operations to make four given numbers equal 24. 
The task requires the language models to combine multiple operations to achieve a target outcome, which can evaluate arithmetic reasoning and logical problem-solving capabilities.
%This task serves as an effective benchmark for evaluating arithmetic reasoning and logical problem-solving capabilities. It requires the language models to combine multiple operations to achieve a target outcome; 
(2) \textit{Mini Crosswords} \cite{NEURIPS2023_271db992}, 
a game of 5×5 mini crosswords and each input includes the 5 horizontal and 5 vertical clues. 
% The expected output is a completed $5×5$ crossword board containing 25 letters. %This task demands both creativity and systematic problem-solving ability. 
To solve this task, the model requires deeper exploration and strategic integration of linguistic clues, allowing us to understand how effectively the evaluated model can expand traditional solution paths and uncover new insights within a complex search space; (3) \textit{Creative Writing} \cite{NEURIPS2023_271db992}, a task to construct a coherent passage with four paragraphs, each ending with one of four given sentences.%, focusing on creativity and planning. 
This task compels LLMs to generate imaginative text with logically sound and contextually rich. Detailed information is provided in Appendix.%utilizing a variety of prompts to gauge the models' innovative and analytical capabilities. 
% To ensure a rigorous and comprehensive assessment, we conduct half of the evaluations using GPT-4~\citep{GPT4}, while the other half are performed by a panel of expert human annotators. In this task, we select 100 open-ended questions from several prompts listed on \textit{Reedsy. com}\footnote{\url{https://blog.reedsy.com/creative-writing-prompts/\#:\~:text=When\%20the\%20idea\%20to\%20start\%20a\%20weekly\%20newsletter}} as our input.

\subsection{Task Performance}
\begin{figure}[htbp]
  \centering
  \begin{subfigure}[b]{0.48\linewidth}
    \centering
    \includegraphics[width=\linewidth]{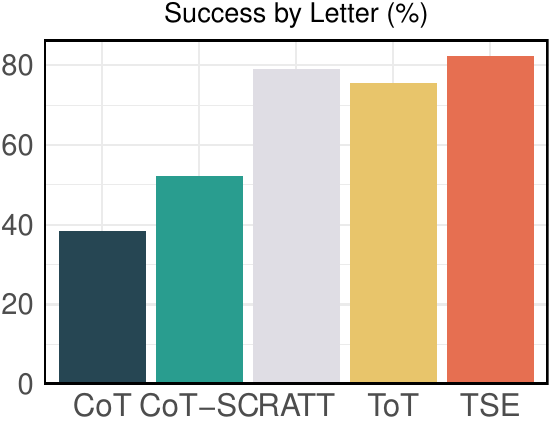}
    \caption{Performance by letter.}
    \label{fig:byletter}
  \end{subfigure}
  \hfill
  \begin{subfigure}[b]{0.48\linewidth}
    \centering
    \includegraphics[width=\linewidth]{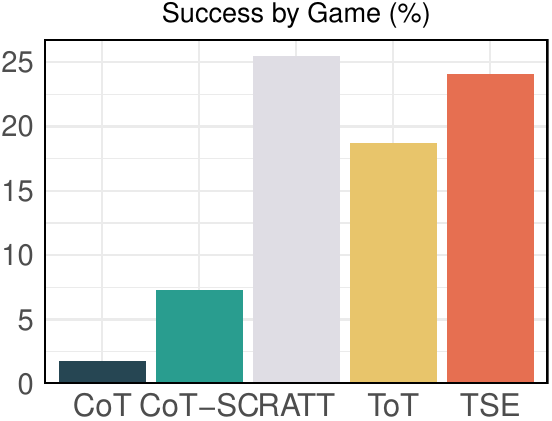}
    \caption{Performance by game.}
    \label{fig:mini}
  \end{subfigure}
  \caption{Performance of GPT-4o-mini on Mini CrossWord.}
  \label{fig:combined}
\end{figure}

The task performance is reported in Table~\ref{tab:overall} and Figure~\ref{fig:combined}. We can see our \model consistently outperforms compared baseline methods across different tasks on most of the metrics.
The effectiveness of \model is attributed to expanding thought structures and exploring solution space to contribute to different aspects of reasoning. Then, we analyze each task separately as follows.

\noindent \textbf{Game of 24.} 
% This task serves as an effective benchmark for evaluating arithmetic reasoning and logical problem-solving capabilities. It requires the language models to combine multiple operations to achieve a target outcome. 
As shown in Table~\ref{tab:overall}, the \model-refined CoT-SC method significantly outperforms other thought structures, including vanilla CoT-SC, with an improvement of $58.56\%$, ToT by $40.50\%$, and RATT by $79.04\%$. Additionally, our enhancement of the basic CoT-SC structure achieves accuracy on GPT-4o-mini that matches the performance of the more complex ToT ($b=5$) implemented on GPT-4~\citep{yao2024tree}. These results highlight the substantial improvements in reasoning capabilities brought about by our method.

% \begin{figure}[h]
%   \centering
%   % \vspace{-2mm}
%   \includegraphics[width=0.92\linewidth]{pic/24-3.pdf}
%   \caption{Comparative performance on the Game of 24.}
%   \label{fig:24}
%   % \vspace{-10pt}
% \end{figure}
% \begin{table}[h]
% \centering
% \resizebox{0.37\textwidth}{!}{
% \begin{tabular}{lc}
% \toprule
% \textbf{Method} & \textbf{Success Rate (\%)} \\
% \midrule
% CoT & 13.3 \\
% CoT-SC (k=5) & 46.7 \\
% ToT (b=5) & 52.7 \\
% RATT & 41.3 \\
% Ours-new & \textbf{64.0} \\
% Ours-combined & \underline{\textbf{74.0}} \\
% \bottomrule
% \end{tabular}
% }
% \caption{Comparative performance on the Game of 24 Task using GPT-4o-mini.}
% \label{tab:comparison_24_task_mini}
% \end{table}

\noindent \textbf{Mini Crossword.}
In this task, we evaluate performance based on two metrics: the proportion of correct letters (out of 25 per game) and the proportion of successfully solved games. As shown in Figure~\ref{fig:mini}, the CoT-SC with \model achieves an impressive accuracy rate of $82.4\%$, significantly outperforming vanilla CoT-SC (by $30.2\%$), surpassing ToT by $6.9\%$. However, for the Game metric, RATT performs slightly better than our method, as it can leverage external knowledge through RAG. Specifically, vanilla CoT-SC suffers from limited exploration of thought paths, which reduces its ability to consistently generate accurate answers. These results show that TSE enhances the model's depth and diversity in exploring potential solutions and generates more accurate and coherent outputs. 
% These results further demonstrate the substantial improvements our method brings to enhance LLMs' reasoning capabilities.
% We select the \textit{Mini Crossword} task because it represents a more complex language-based reasoning problem that demands both creativity and a systematic approach to problem-solving, making it well-suited for evaluating the comprehensive reasoning abilities of our method. Unlike tasks such as the Game of 24, which involve relatively shallow reasoning paths, the \textit{Mini Crossword} requires deeper exploration and strategic integration of linguistic clues, making it ideal for testing the enhanced exploration and thought diversity enabled by our approach. This task allows us to evaluate how effectively \model can expand beyond traditional solution paths and uncover new insights within a more challenging search space.

% \begin{figure}[h]
%   \centering
%   % \vspace{-2mm}
%   \includegraphics[width=1.02\linewidth]{pic/cross-2.pdf}
%   \caption{Performance of different methods on GPT-4 for task Mini Crossword.}
%   \label{fig:cross}
%   % \vspace{-3mm}
% \end{figure}

% \begin{table}[h]
% \centering
% \resizebox{0.3\textwidth}{!}{%
% \begin{tabular}{lcc}
% \toprule
% \textbf{Method} & \textbf{Letter (\%)} & \textbf{Game (\%)} \\
% \midrule
% CoT & 38.4 & 1.8 \\
% CoT-SC & 52.2 & 7.3 \\
% ToT & 75.5 & 18.7 \\
% RATT & 79.2 & 25.5 \\
% Ours & \textbf{82.4} & \textbf{24.1} \\
% \bottomrule
% \end{tabular}
% }
% \caption{Comparative Performance of Methods in Different Categories}
% \label{tab:performance_categories}
% \end{table}

\noindent \textbf{Creative Writing.} 
% \begin{figure}[h]
%   \centering
%   % \vspace{-2mm}
%   \includegraphics[width=1\linewidth]{pic/creative-2.pdf}
%   \caption{Performance of different methods on GPT-4o-mini for task Creative Writing.}
%   \label{fig:writing}
%   % \vspace{-3mm}
% \end{figure}
As shown in Table~\ref{tab:overall}, the evaluation criteria included \textit{Soundness}, \textit{Innovation}, \textit{Content Coherence of Reasoning}, and \textit{Clarity of Expression}, applying consistently across all evaluations. We rate each metric on a scale from 0 to 10 in 0.5-point increments, and calculate the overall performance score as the average of these four dimensions. A detailed figure can be found in the Appendix.

\subsection{Ablation Study}

% \begin{table}[htbp]
%   \centering
%   \caption{Comparative Analysis of Node Selection Methods Across Multiple Metrics}
%   \label{tab:2-connect}
%   \resizebox{0.48\textwidth}{!}{%
%     \begin{tabular}{c|c|c|c|c|c|c}
%       \toprule
%       \hline
%       \multicolumn{2}{c|}{\textbf{Game of 24}} & \multicolumn{5}{c}{\textbf{Creative Writing}} \\ \hline
%       \textbf{Method} & \textbf{Success Rate (\%)} & \textbf{Soundness} & \textbf{Innovation} & \textbf{Coherence} & \textbf{Expression} & \textbf{Overall} \\ \hline
%       Original & 5.33 & 4.73 & 4.52 & 4.85 & 4.62 & 4.68 \\ \hline
%       Random & 3.95 & 4.7 & 4.77 & 4.51 & 4.18 & 4.54 \\ \hline
%       TSE & 17.11 & 5.05 & 5.45 & 5.05 & 4.8 & 5.09 \\ \hline
%     \end{tabular}%
%   }
% \end{table}

\paragraph{Key Node Selection.} To validate the importance of key node selection, we compare (1) the original CoT-SC, (2) the random selection of one node per chain for generation, and (3) our key node selection method. As shown in Table~\ref{tab:combined_methods}, our method significantly improves on \textit{Game of 24} compared to original and random approaches. In \textit{Creative Writing}, particularly for innovation, our method also demonstrates notable enhancements, proving its effectiveness in selecting nodes for generation. The superior performance of our key node selection method is primarily due to its ability to identify nodes containing crucial information for problem-solving. The model only utilizes the most influential information and thus avoids redundancy, thus enhancing the model's efficiency and creativity.

\paragraph{Connection Node Selection.} We then test the necessity of connection node selection by comparing (1) randomly selecting key nodes to connect, (2) selecting the key nodes based on lower inference layers, and (3) our method. As shown in Table~\ref{tab:combined_methods}, our method is significantly superior to random selection and layer-based selection on overall reasoning performance, particularly in coherence. This is because random selection lacks specificity and cannot guarantee that the chosen nodes will effectively support the content of new nodes. Although layer-based selection considers the structural hierarchy of information, it does not necessarily reflect the actual importance or applicability of the information. Our method analyzes relationships and reasoning paths between nodes, thus maintaining and strengthening the information flow's coherence and depth.

\paragraph{Collaborative Reasoning.} In this experiment, we test the importance of the collaboration method in \model. We compare four collaboration methods: (1) aggregating answers through majority voting, (2) randomly sampling partial thought chains, (3) using only the outcomes from newly generated chains, and (4) implementing the \model collaborative method. As shown in Table~\ref{tab:combined_methods}, \model consistently outperforms other methods in both tasks. Meanwhile, using only new chains also achieves a high success rate in \textit{Game of 24}, demonstrating their potential to offer innovative solutions. However, since these new chains lack integration with original chain information, their performance is still weaker than that of \model's comprehensive collaboration methods. In \textit{Creative Writing}, using only new chains proved better than majority voting. This suggests that while new chains can offer innovative content, they may lack coherence and depth without sufficient integration with the original chains.

\begin{table*}[htbp]
  \centering
  \small
  % \vspace{-2mm}
  % \resizebox{0.88\textwidth}{!}{%
    \begin{tabular}{c|ccc|ccc}
      \toprule
      \midrule
      \textbf{Method} & \multicolumn{3}{c|}{\textbf{GPT}} & \multicolumn{3}{c}{\textbf{Llama}} \\
      \hline
      & \textbf{Success Rate (\%)} & \textbf{Overall} & \textbf{Coherence} & \textbf{Success Rate (\%)} & \textbf{Overall} & \textbf{Coherence} \\
      \midrule
      
      \textbf{Key Node Selection} \\
      \midrule
      Original        & 13.3 & 5.12 & 5.03 & 5.33  & 4.68 & 4.85 \\
      Random          & 51.5 & 5.20 & 4.97 & 3.95  & 4.54 & 4.51 \\
      TSE             & 74.0 & 6.03 & 5.74 & 18
      75 & 5.09 & 5.05 \\
      \midrule
      \textbf{Connection Node Selection} \\
      \midrule
      Random          & 62.8 & 5.91 & 5.66 & 9.52  & 4.82 & 4.85 \\
      Layer-based     & 67.3 & 5.95 & 5.69 & 6.12  & 4.84 & 4.88 \\
      TSE             & 74.0 & 6.03 & 5.74 & 18.75 & 5.09 & 5.05 \\
      \midrule
      \textbf{Collaborative Reasoning} \\
      \midrule
      Majority-Vote   & 46.7 & 5.81 & 5.87 & 10.0  & 4.85 & 4.85 \\
      Random Sampling & 27.3 & 5.90 & 5.89 & 12.5  & 4.92 & 4.90 \\
      New Chains Only & 64.0 & 5.95 & 5.87 & 18.06 & 5.11 & 5.06 \\
      TSE             & 74.0 & 6.03 & 5.90 & 18.75 & 5.09 & 5.05 \\
      \hline
      \bottomrule
    \end{tabular}
  % }
  \caption{Impact of node selection methods.}
  \label{tab:tse_impact}
  \vspace{-2ex}
\end{table*}

\subsection{Strategy Analysis}
\begin{figure}[htbp]
  \centering
  % \vspace{-2mm}
  \includegraphics[width=1.01\linewidth]{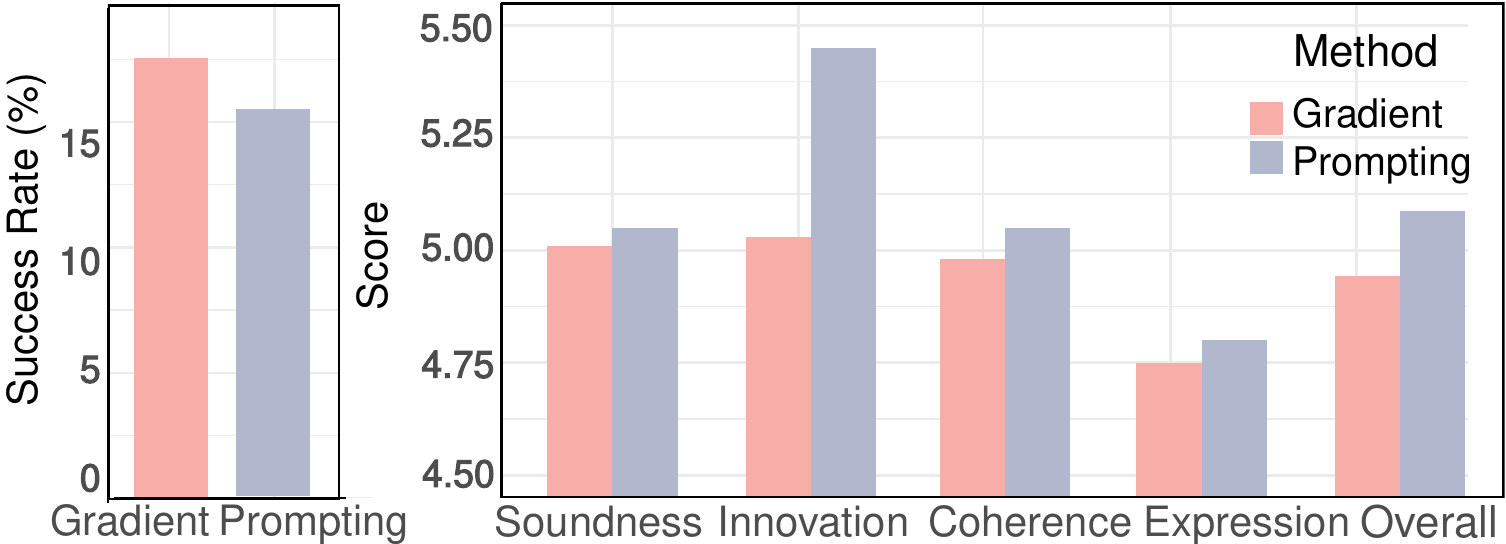}
  \caption{Performance of key node selection methods.}
  \label{fig:2-keynode}
  % \vspace{-3mm}
\end{figure}

\paragraph{Impact of Key Node Selection Methods.} 
In this experiment, we compare the impact of two key node selection methods—gradient-based and self-prompting—on Llama's CoT-SC reasoning performance. As shown in Figure~\ref{fig:2-keynode}, both methods achieve comparable results in the \textit{Game of 24}, while self-prompting performs better in \textit{Creative Writing}. This indicates both approaches effectively select critical nodes, but self-prompting additionally helps the model revisit overlooked thought connections and blind spots. Consequently, the model generates more reliable information and produces more creative content. Thus, for tasks demanding higher creativity, self-prompting can encourage the model to move beyond conventional thought patterns and foster innovative solutions.
% In this experiment, we compare the impact of our two key node selection methods on model reasoning performance. Here we contrast the gradient-based and self-prompting selection on Llama's CoT-SC structures. As shown in Figure~\ref{fig:2-keynode}, both methods perform similarly in task 
% \textit{Game of 24}, while in \textit{Creative Writing}, the self-prompting method shows superior results. This suggests that both methods are capable of selecting appropriate key nodes. Meanwhile, the self-prompting methods guide the model to review known information as well as consider how to connect different thought fragments. As both methods select the key node of the chain, self-promoting explores blind spots that the model might overlook in the usual autoregressive generation process. By delving deeper into these spaces, the model generates more reliable information from the selected nodes and provides more creative content when generating new nodes. Thus, for tasks requiring high creativity and diverse thinking, a potential strategy is to prompt the model to break free from conventional patterns and generate innovative insights.


\paragraph{Comparison of Connection Node Selection Methods.} We then compare gradient comparison and semantic relevance for selecting connection nodes when generating new reasoning paths on Llama. As shown in Table~\ref{tab:combined_methods}, semantic relevance selection achieves better performance on both \textit{Game of 24} and coherence in \textit{Creative Writing}. The difference possibly arises from two main reasons: (1) Nodes with larger relative gradients do not necessarily have a stronger influence on the content when generating new nodes. (2) Compared to gradient-based selection, the semantic relevance method more effectively captures the actual content and meaning relationships between nodes, leading to more suitable connection points for new nodes. This demonstrates that considering semantic relevance when selecting connection points provides more effective support for reasoning performance.

\paragraph{Comparison of Collaborative Reasoning Methods.} In this experiment, we compare the performance of CWS and LM-as-a-Judge. As shown in Table~\ref{tab:combined_methods}, in task \textit{Game of 24}, which requires dense logic and has well-defined objectives, the LM-as-a-Judge method accurately and quickly identifies the correct answers by directly evaluating each thought chain and selecting the optimal solution. This method relies on the model's ability to assess individual outputs, allowing for rapid selection of the best solution—especially useful in tasks that require precise calculations and quick responses. In task \textit{Creative Writing}, which requires innovation and diversity, CWS evaluates and weights the contributions of each thought chain. The final output represents a blend of multiple perspectives and enhances the richness and depth of the narrative significantly. We further analyze the efficacy and provide a case study to illustrate the effectiveness of \model generation and collaboration below.

\begin{table}
  \centering
  \scriptsize
  \resizebox{\linewidth}{!}{%
    \begin{tabular}{c|c|c|c}
      \toprule
      \hline
      \multicolumn{2}{c|}{\textbf{Game of 24}} & \multicolumn{2}{c}{\textbf{Creative Writing}} \\ 
      \hline
      \textbf{Method} & \textbf{Success Rate (\%)} & \textbf{Overall} & \textbf{Coherence} \\
      \hline
      \multicolumn{4}{c}{Connection Node Selection Methods} \\
      \hline
      Gradient & 13.61 & 4.88 & 4.86 \\
      Relevance & 16.32 & 4.91 & 4.94 \\ 
      \hline
      \multicolumn{4}{c}{Collaborative Reasoning Methods} \\ \midrule
      CWS & 12.93 & 4.97 & 5.01 \\ 
      LM-as-a-Judge & 18.05 & 5.09 & 5.05 \\ 
      \bottomrule
    \end{tabular}%
  }
  \caption{Comparison of connection node selection and collaborative reasoning strategy.}
  \label{tab:combined_methods}
\end{table}

\section{Experimental Settings}

\subsection{Baselines} %To demonstrate the performance improvements brought by our method, 
\label{subsec:baselines}
We apply our method to the simplest multi-chain thought structure and compare the results with several baseline methods, including CoT, Vanilla CoT-SC, ToT, and RATT. This comparison aims to illustrate how our method enhances the thought structure compared to existing approaches. For a consistent evaluation, we standardize the depth to five across all methods. Specifically, 
we set the number of chains to $k=5$ and generate $b=5$ candidates at each step for CoT-SC and ToT, respectively. 
For RATT, the model generates and integrates five candidate results at each decision point and uses \texttt{Wikipedia}%~\citep{WikipediaMainPage} 
\footnote{\url{https://en.wikipedia.org}} as the external resource.

\subsection{Task Description} 
\label{subsec:taskdescription}
We evaluate \model and the baseline methods on three reasoning datasets with specific tasks: (1) \textit{Game of 24}, a mathematical challenge whose objective is to use the four basic arithmetic operations to make four given numbers equal 24. 
The task requires the language models to combine multiple operations to achieve a target outcome, which can evaluate arithmetic reasoning and logical problem-solving capabilities.
%This task serves as an effective benchmark for evaluating arithmetic reasoning and logical problem-solving capabilities. It requires the language models to combine multiple operations to achieve a target outcome; 
(2) \textit{Mini Crosswords}, 
a game of 5×5 mini crosswords and each input includes the 5 horizontal and 5 vertical clues. The expected output is a completed $5×5$ crossword board containing 25 letters. %This task demands both creativity and systematic problem-solving ability. 
To solve this task, the model requires deeper exploration and strategic integration of linguistic clues, allowing us to understand how effectively the evaluated model can expand traditional solution paths and uncover new insights within a complex search space; (3) \textit{Creative Writing}, a task to construct a coherent passage with four paragraphs, each ending with one of four given sentences.%, focusing on creativity and planning. 
This task compels LLMs to generate imaginative text withlogically sound, and contextually rich. %utilizing a variety of prompts to gauge the models' innovative and analytical capabilities. 
To ensure a rigorous and comprehensive assessment, we conduct half of the evaluations using GPT-4~\citep{GPT4}, while the other half are performed by a panel of expert human annotators. In this task, we select 100 open-ended questions from several prompts listed on \textit{Reedsy. com}\footnote{\url{https://blog.reedsy.com/creative-writing-prompts/\#:\~:text=When\%20the\%20idea\%20to\%20start\%20a\%20weekly\%20newsletter}} as our input.

\subsection{Thought Chain Efficacy}
\label{subsec:Thoughtchain}
In this experiment, we conduct a detailed comparison between the success rates of original thought chains and new chains generated by our method in the \textit{Game of 24} task. Experimental results indicate that for each problem, the success rate of having at least one correct answer in the original CoT-SC is approximately $50.7\%$, while the new chains generated by our method achieve a success rate of $64.0\%$. Moreover, despite improving accuracy, the overlap of problems successfully solved by both the old and new chains is only $35.3\%$. This low level of overlap suggests that our method effectively explores new areas that the original thought structures do not address and brings a significant increase in the overall success rate.

\subsection{Creative Writing\label{subsec:crea}}
\begin{figure}[h]
  \centering
  % \vspace{-2mm}
  \includegraphics[width=\linewidth]{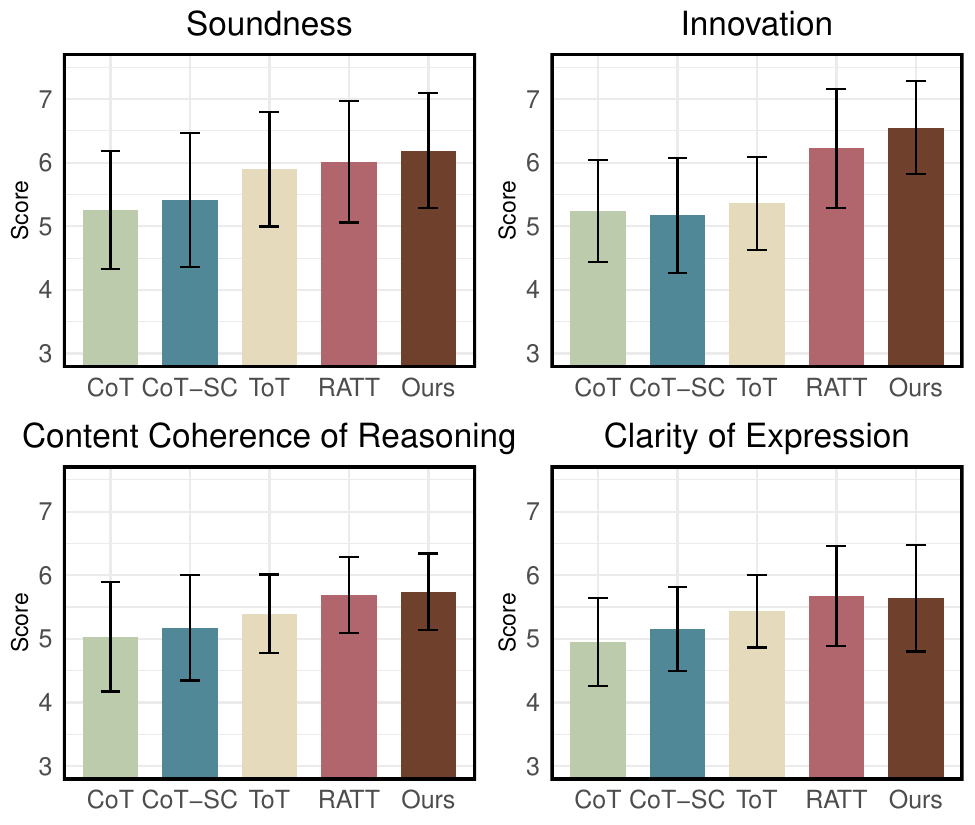}
  \caption{Performance of different methods on GPT-4o-mini for task Creative Writing.}
  \label{fig:writing}
  % \vspace{-3mm}
\end{figure}

In Figure~\ref{fig:writing} we provide detailed results on task \textit{Creative Writing}. \model method outperforms other methods in three aspects and is only slightly weaker than RATT with RAG.

\section{Detailed Information of Case Study}
\label{sec:case}
\subsection{Case Analysis}
\begin{figure*}[htbp]
  \centering
  % \vspace{-2mm}
  \includegraphics[width=1.0\linewidth]{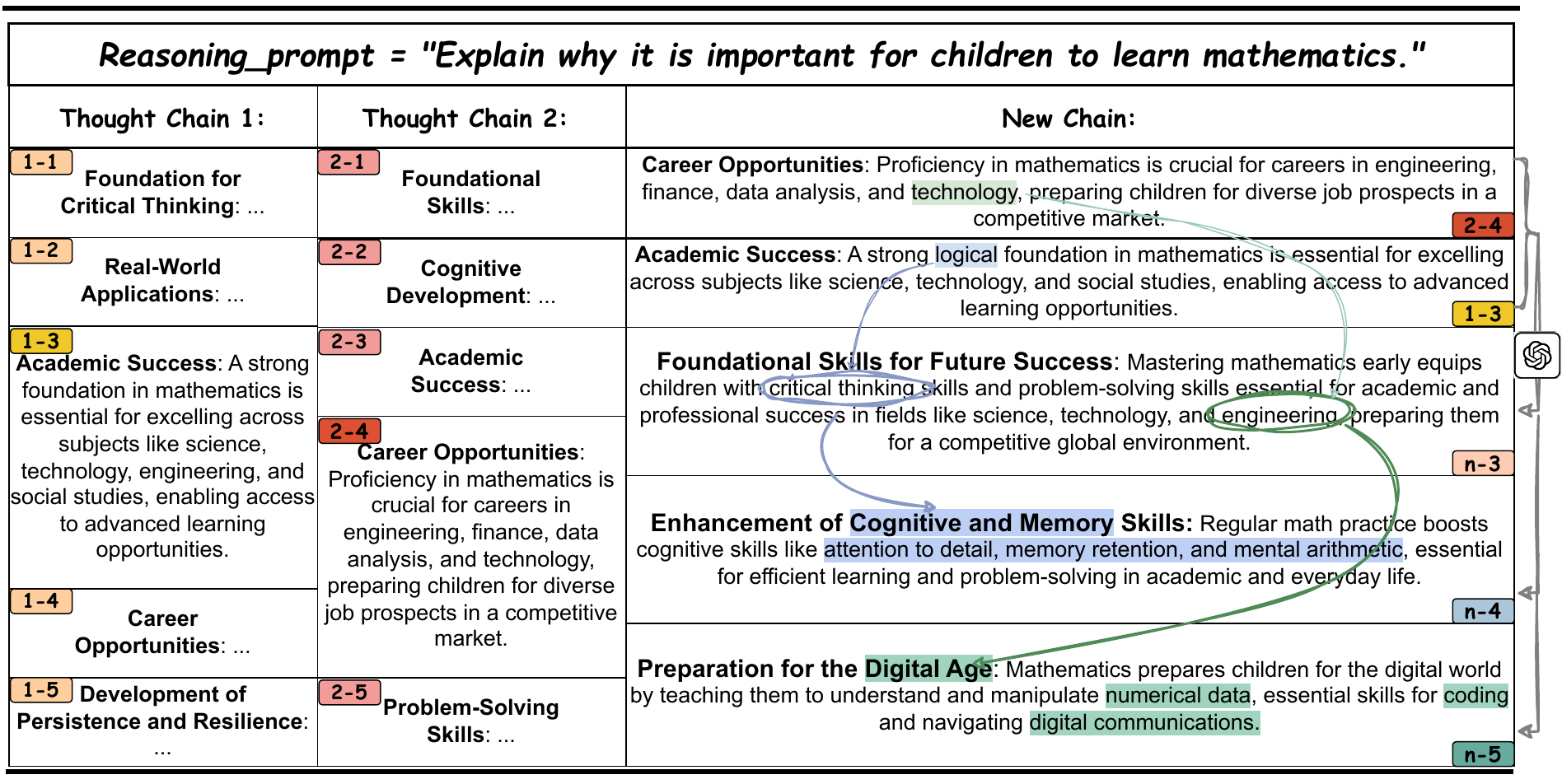}
  \caption{Case study of comparison of original and new chains.}
  \label{fig:case}
  % \vspace{-3mm}
\end{figure*}

As our approach significantly enhances the reasoning capabilities of thought structures while maintaining minimal computational cost, it effectively addresses gaps within existing structures. Figure~\ref{fig:case} and Table~\ref{tab:case_math} provides a specific example of this enhancement by comparing the original thought chains with the new chains generated by our method. In the provided case study, we see the reasoning prompt \textit{``Explain why it is important for children to learn mathematics''}. The original thought chains (Thought Chain 1 and Thought Chain 2) are linear and somewhat limited in scope, focusing and repeating on foundational aspects like \textit{``Critical Thinking'', ``Academic Success''} and \textit{``Career Opportunities''}. However, the new chain generated by our method explores additional dimensions, such as `\textit{`Enhancement of Cognitive and Memory Skills''} and \textit{``Preparation for the Digital Age''}. These new perspectives enrich the argument and provide a more comprehensive understanding of the importance of mathematics. 

\begin{table*}[ht]
\centering
\small
\begin{tabular}{|p{0.3\linewidth}|p{0.3\linewidth}|p{0.3\linewidth}|}
\hline
\textbf{Thought Chain 1} & \textbf{Thought Chain 2} & \textbf{New Chain} \\ \hline
\textbf{Foundation for Critical Thinking:} Mathematics fosters critical and logical thinking, enhancing problem-solving and analytical skills necessary for daily life. & \textbf{Foundational Skills:} Mathematics equips children with essential daily skills like counting, measuring, and problem-solving, crucial for tasks such as budgeting, shopping, and time management. & \textbf{Career Opportunities:} Proficiency in mathematics is crucial for careers in engineering, finance, data analysis, and technology, preparing children for diverse job prospects in a competitive market. \\ \hline
\textbf{Real-World Applications:} Mathematics is crucial for managing daily activities like finances, cooking, and scheduling, empowering children with the confidence to handle them effectively. & \textbf{Cognitive Development:} Math enhances logical reasoning, critical thinking, and problem-solving abilities, benefiting other subjects and real-life situations. & \textbf{Academic Success:} A strong foundation in mathematics is essential for excelling across subjects like science, technology, engineering, and social studies, enabling access to advanced learning opportunities. \\ \hline
\textbf{Academic Success:} A strong mathematics foundation is essential for excelling in subjects like science, technology, engineering, and social studies, paving the way for advanced educational opportunities. & \textbf{Academic Success:} Proficiency in mathematics underpins success in STEM fields, crucial for today's job market, and opens doors to advanced educational opportunities. & \textbf{Foundational Skills for Future Success:} Mastering mathematics early equips children with critical thinking and problem-solving skills essential for academic and professional success in fields like science, technology, and engineering, preparing them for a competitive global environment. \\ \hline
\textbf{Career Opportunities:} Proficiency in mathematics is key for careers in engineering, finance, technology, medicine, and data science, equipping children for diverse job prospects in a competitive market. & \textbf{Career Opportunities:} Proficiency in mathematics is crucial for careers in engineering, finance, data analysis, and technology, preparing children for a broad range of opportunities in a competitive job market. & \textbf{Enhancement of Cognitive and Memory Skills:} Regular math practice boosts cognitive skills like attention to detail, memory retention, and mental arithmetic, essential for efficient learning and problem-solving in academic and everyday life. \\ \hline
\textbf{Development of Persistence and Resilience:} Mathematics teaches children systematic problem-solving, a skill invaluable in academics and beyond. & \textbf{Problem-Solving Skills:} Mathematics teaches children systematic problem-solving, a skill invaluable in academics and beyond. & \textbf{Preparation for the Digital Age:} Mathematics prepares children for the digital world by teaching them to understand and manipulate numerical data, essential skills for coding and navigating technological advancements. \\ \hline
\end{tabular}
\caption{Reasoning Prompt: Explain why it is important for children to learn mathematics.}
\label{tab:case_math}
\end{table*}

\clearpage

\end{document}